\documentclass[preprint,12pt,authoryear]{elsarticle}

\usepackage{amssymb}

\usepackage{graphicx}
\graphicspath{ {images/} }
\usepackage{caption}     
\usepackage{subcaption}
\usepackage{booktabs}
\usepackage{multirow}
\usepackage{array}
\usepackage{makecell}
\usepackage{adjustbox}

\newcolumntype{M}[1]{>{\centering\arraybackslash}m{#1}}

\usepackage{amsthm}
\usepackage{amsmath}
\usepackage{url}
\usepackage{algorithm}
\usepackage{algpseudocode}
\usepackage{algorithmicx}
\usepackage{lineno}
\usepackage{xcolor}
\usepackage{hyperref}

\algdef{SE}[DOWHILE]{Do}{doWhile}{\algorithmicdo}[1]{\algorithmicwhile\ #1}%
\algrenewcommand\algorithmicrequire{\textbf{Input:}}
\algrenewcommand\algorithmicensure{\textbf{Output:}}

\journal{Automation in Construction}

\begin{document}

\begin{frontmatter}

\title{InfraDiffusion: zero-shot depth map restoration with diffusion models and prompted segmentation from sparse infrastructure point clouds}

\author[a]{Yixiong Jing}
\author[b]{Cheng Zhang}
\author[a]{Haibing Wu *}
\author[a]{Guangming Wang}
\author[a]{Olaf Wysocki}
\author[a]{Brian Sheil}

\affiliation[a]{
    organization={Construction Engineering, University of Cambridge},
    addressline={Trumpington Street}, 
    city={Cambridge},
    postcode={CB2 1PZ}, 
    state={Cambridge},
    country={UK}
}

\affiliation[b]{
    organization={College of Civil Engineering, Hunan University},
    addressline={Yuelu South Road}, 
    city={Changsha},
    postcode={410082}, 
    state={Hunan Province},
    country={China}
}
    
\begin{abstract}

Point clouds are widely used for infrastructure monitoring by providing geometric information, where segmentation is required for downstream tasks such as defect detection. Existing research has automated semantic segmentation of structural components, while brick-level segmentation (identifying defects such as spalling and mortar loss) has been primarily conducted from RGB images. However, acquiring high-resolution images is impractical in low-light environments like masonry tunnels. Point clouds, though robust to dim lighting, are typically unstructured, sparse, and noisy, limiting fine-grained segmentation. We present InfraDiffusion, a zero-shot framework that projects masonry point clouds into depth maps using virtual cameras and restores them by adapting the Denoising Diffusion Null-space Model (DDNM). Without task-specific training, InfraDiffusion enhances visual clarity and geometric consistency of depth maps. Experiments on masonry bridge and tunnel point cloud datasets show significant improvements in brick-level segmentation using the Segment Anything Model (SAM), underscoring its potential for automated inspection of masonry assets. Our code and data is available at \url{https://github.com/Jingyixiong/InfraDiffusion-official-implement}.

\end{abstract}

\begin{highlights}
\item Introduce a zero-shot framework, InfraDiffusion, to restore depth maps from masonry point clouds.
\item Propose a virtual camera projection for depth map generation from point clouds.
\item Adapt DDNM with boundary masking for depth image restorations using pre-trained diffusion models.
\item Improve segmentation metrics across five datasets using SAM segmentation.
\end{highlights}


\end{frontmatter}

\section{Introduction}\label{sec1}

Masonry infrastructure forms a significant part of the transport network and civil engineering assets in the UK \citep{orban2004assessment}. These structures require regular inspection to ensure long-term safety and functionality \citep{acikgoz2018sensing}. However, these inspections are often time-consuming and reliant on visual judgment of engineers, which introduces both subjectivity and variability into the assessment process \citep{brackenbury_2022}.

LiDAR scan point clouds are increasingly adopted to support infrastructure assessment, enabling efficient acquisition of the as-is geometry of built environments \citep{wang2015smart, dai2013three, lubowiecka2009historic, shanoer2018evaluate}. To perform structural analysis from the point cloud, segmentation is required as a pre-processing step to identify structural elements. Recent advances in deep learning (DL) have enabled the automated segmentation of structural `components' (such as arches, spandrels, piers, or walls) in masonry bridge point clouds \citep{jing2022segmentation, jing2024lightweight}. This component-level segmentation enables a range of downstream tasks, including geometric modelling \citep{jing2023method, han2025automatic}, deformation tracking \citep{ye2018mapping, acikgoz2017evaluation}, and defect detection \citep{jing4819836anomaly, han2025automatic}.

Whilst previous work has predominantly focused on component-level segmentation, finer segmentation of individual bricks (i.e., “brick-level segmentation”) from point clouds remains unexplored. Brick-level segmentation is important for automating early-stage defect detection \cite{dais2021automatic, loverdos2022automatic, hallee2021crack, ye2024sam}, particularly in large-scale masonry structures. Different from steel and concrete structures, masonry structures do not behave as continuous materials but as assemblages of discrete units where the brick–mortar interaction governs both stiffness and failure mechanisms \citep{Loureno2013ComputationalSF, lemos2007discrete}. Different deterioration mechanisms can be introduced at this scale: mortar joints typically experience erosion and volume loss, while bricks may undergo spalling, cracking, and surface detachment \citep{giordano2002modelling, milani2006homogenised}. Identifying individual bricks within point clouds is therefore crucial for distinguishing between brick- and mortar-related damage, enabling more accurate structural diagnosis \citep{wu2019concrete, katsigiannis2023deep} and supporting advanced analysis approaches such as discrete element modelling \citep{lemos2007discrete}.

Previous studies \citep{dais2021automatic, ye2024sam} have typically performed brick-level segmentation on images, which provide dense visual information for annotation. However, acquiring high-quality images is often impractical in low-light conditions common in tunnels \citep{cheng2019automatic}, indoor facilities \citep{luo2023indoor}, and underground pipes \citep{you2025construction}. In contrast, point clouds collected by active sensors such as LiDAR are robust to poor lighting \citep{qu2014challenges}, but their sparsity and noise significantly limit fine-grained segmentation. These limitations highlight the need for methods that enable brick-level segmentation directly from point clouds.

However, it is nontrivial to automate segmentation directly on point clouds using DL models due to two significant drawbacks:
\begin{enumerate}[(i)]
    \item \textbf{Lack of large annotated datasets:} 3D DL models designed for point cloud segmentation typically require large and well-annotated datasets for extensive training, which are time-consuming and costly to produce.
    
    \item \textbf{Lack of generalisation:} Unlike 2D vision \citep{kirillov2023segment} and NLP \citep{guo2025deepseek}, the point cloud domain lacks robust foundation models. As a result, existing 3D DL models trained on narrow datasets often fail to generalise across different types of infrastructure.
\end{enumerate}

Unlike 2D vision \citep{kirillov2023segment} and NLP \citep{guo2025deepseek}, the point cloud domain lacks robust foundation models. As a result, existing 3D DL models trained on narrow datasets often fail to generalise across different types of infrastructure.

To overcome these limitations, recent studies have proposed project-based segmentation methods, which project point clouds into structured 2D depth maps \citep{zhang2022unrollingnet, chen2024combining}. The 2D depth maps can leverage powerful image-based foundation models that provide greater robustness and generalisation across diverse scenes and segmentation tasks. For example, \citet{ye2025sam4tun} recently demonstrated the use of the Segment Anything Model (SAM) \citep{kirillov2023segment} to perform zero-shot instance segmentation of tunnel lining segments, which achieves accurate and robust results without training.

However, the generalisation of current projection-based methods remains limited due to two challenges:
\begin{enumerate}[(i)]
    \item These methods have been predominantly applied to tunnel environments, where cylindrical projection is used for unwrapping the geometry into depth images. Cylindrical projection does not generalise well to other types of infrastructure with more complex or irregular topologies, such as masonry bridges.
    
    \item The quality of the resulting depth maps heavily depends on the equipment noise, registration errors, and density of the original point cloud. Therefore, the projected depth maps are always incomplete and noisy in real-world scenarios, which makes them unsuitable for brick-level segmentation tasks.
\end{enumerate}

To address challenge (i), the autonomous driving domain has introduced virtual camera projection to allow flexible viewpoint selection by simulating the optics of real cameras and adapting to any complex 3D scene. This approach has been applied across a range of tasks, including segmentation \citep{wu2019squeezesegv2, lyu2020learning, krawciw2024lasersam}, 3D object detection \citep{chen2017multi, meyer2019lasernet}, and multi-view data fusion \citep{massa2024adapting}. Unlike cylindrical or planar projections \citep{chen2024combining, ye2025sam4tun}, virtual camera projection enables depth maps generation for infrastructure with irregular or complex forms (such as masonry bridges), where global unfolding techniques often fail to preserve structural continuity. 

To address challenge (ii) related to noisy and sparse depth maps, image restoration (IR) technologies provide a feasible solution for enhancing image quality. Researchers have explored DL models to guide the restoration process, which can be classified into non-generative, GAN-based, and diffusion-based methods. Non-generative methods explicitly estimate the degradation parameters of an image (such as noise levels \citep{chen2019real}, blur kernels for super-resolution \citep{dong2014learning, dong2015image}, or missing regions for inpainting \citep{cao2022learning}) and then restore high-quality images using the predicted information. However, these models often struggle to generalise to complex real-world degradation patterns due to oversimplified degradation assumptions. To address this, Generative Adversarial Networks (GANs) have been used to implicitly learn both the underlying data distribution and the degradation process, achieving promising results in tasks such as super-resolution \citep{fritsche2019frequency, wang2021unsupervised, wang2021real}. Nonetheless, GAN-based methods are limited by training stability due to the use of dual-network architectures and intricate adversarial loss functions.

Recent advancements in Denoising Diffusion Probabilistic Models (DDPMs) have shown remarkable performance in synthesising visually compelling images \citep{ho2020denoising, song2020denoising, rombach2022high}. Leveraging these generative strengths, researchers have extended DDPMs to IR tasks, either by fine-tuning pre-trained DDPMs \citep{lin2024diffbir, saharia2022image, zhu2024flowie} or by training new DDPMs from scratch \citep{wang2023exposurediffusion, lo2024roofdiffusion}. However, most of the existing diffusion-based methods have focused on restoring RGB images, with limited attention on depth maps or geometric modalities. Recently, \cite{lo2024roofdiffusion} proposed RoofDiffusion to effectively restore roof height maps from severely incomplete depth maps by conditioning on building footprints to address sparsity and occlusion. While DDPMs achieve robust and effective IR tasks, they require paired training data comprising clean and corrupted images as additional conditions in the loss functions. This reliance limits their applicability to infrastructure point clouds, where ground-truth (GT) depth maps are rarely available due to the scarcity of dense geometric information. 
\begin{figure*}[t!]
    \centering
    \includegraphics[width=\linewidth]{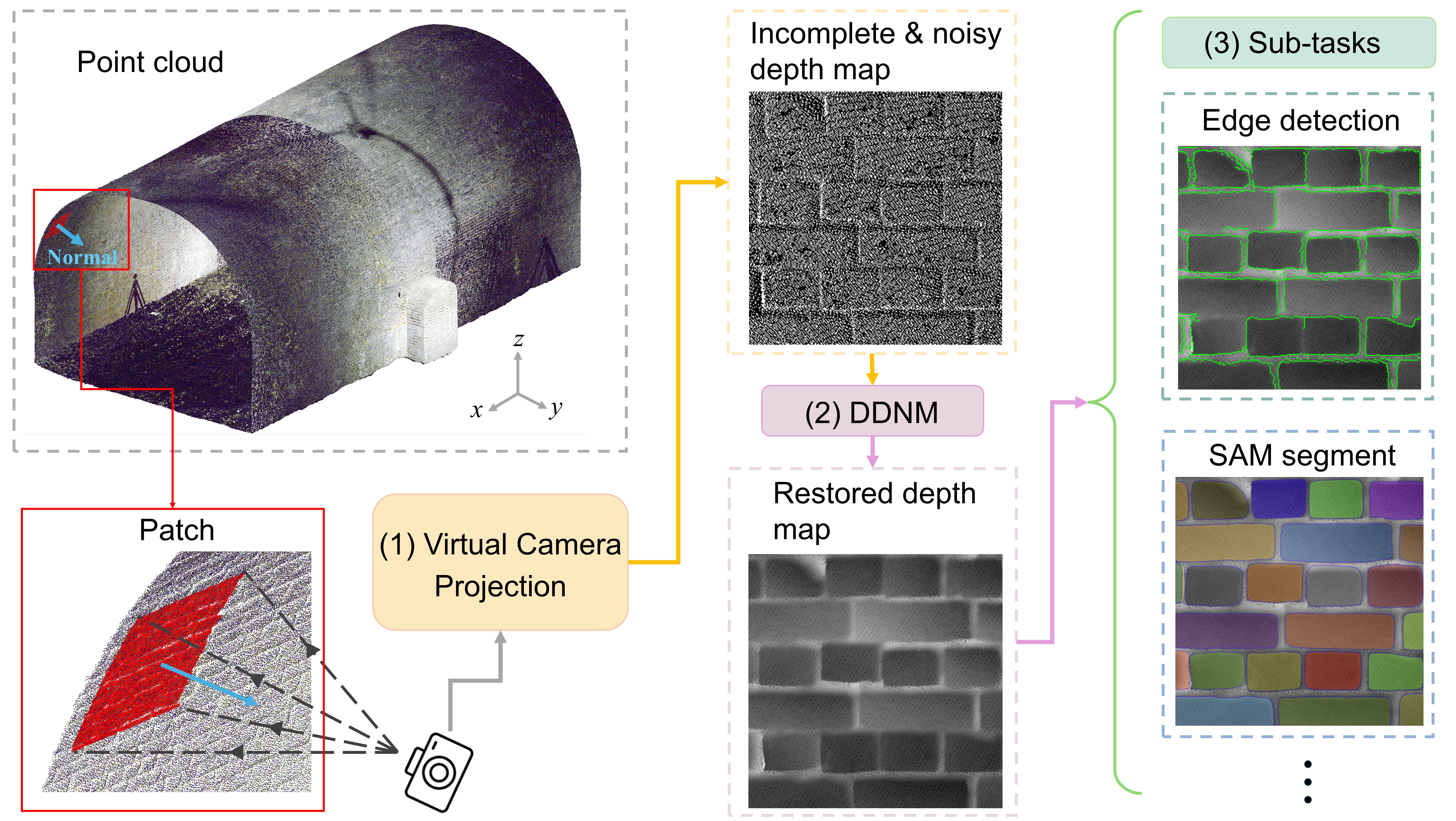}
    \caption{Overview of the InfraDiffusion pipeline.}
    \label{fig: overview_pipeline}
\end{figure*}

To overcome the need for GT depth maps, we adapt the Denoising Diffusion Null-space Model (DDNM) \citep{wang2022zero}, a zero-shot IR framework that uses pre-trained DDPMs to provide strong image priors. DDNM decomposes the restoration process into range-space and null-space components, ensuring strict data consistency by fixing the known degraded part and refining only the unknown null-space through sampling images from pre-trained DDPMs. This approach eliminates the need for task-specific training and enables effective restoration in the absence of clean depth maps. In our work, we adapt DDNM to restore sparse and noisy depth maps projected from infrastructure point clouds to enable intricate downstream tasks such as brick-level segmentation. 

To overcome the need for GT depth maps, we adapt the Denoising Diffusion Null-space Model (DDNM) \citep{wang2022zero}, a zero-shot image restoration framework that leverages pre-trained DDPMs as strong image priors. DDNM decomposes the restoration process into range-space and null-space components, ensuring strict data consistency by fixing the known degraded part and refining only the unknown null-space through sampling from pre-trained DDPMs. In our work, we extend DDNM by introducing boundary masks that constrain the diffusion generation to regions where point cloud projections are valid. Our modification addresses a common challenge in utilising virtual camera projection in infrastructure point cloud datasets, where virtual camera viewpoints often lead to incomplete depth maps near the boundaries of point clouds.

By preventing spurious generation in empty regions, we introduce a complete zero-shot pipeline, i.e., InfraDiffusion, that restores sparse and noisy depth maps projected from infrastructure point clouds without task-specific training, thereby enabling fine-grained tasks such as brick-level segmentation. As illustrated in Figure~\ref{fig: overview_pipeline}, the pipeline begins by extracting surface patches from the point cloud and projecting them into 2D depth maps using virtual cameras. These incomplete and noisy depth maps are then restored using InfraDiffusion, which enhances depth quality without requiring GT images and fine-tuning. We validate InfraDiffusion on two representative datasets: a masonry tunnel with three different sampled sections and three masonry bridges, each presenting different challenges in terms of point cloud sparsity and geometric complexity. To evaluate the effectiveness of the restoration, we perform brick-level semantic segmentation using the Segment Anything Model (SAM) \citep{ye2024sam} with coordinate prompts to isolate individual bricks. The results demonstrate that the restored depth maps significantly improve segmentation accuracy. As a fully zero-shot framework, InfraDiffusion offers a scalable and adaptable solution for fine-grained geometric analysis of infrastructure point clouds in real-world inspection workflows.

The main contributions of this work can be summarised as follows:
\begin{enumerate}[(i)]
    \item We propose InfraDiffusion, a zero-shot pipeline that leverages virtual camera projection and image restoration to improve the quality of depth maps derived from masonry infrastructure point clouds.

    \item We extend the DDNM by introducing boundary masks that constrain generation to valid projection regions, addressing boundary effects common in infrastructure point clouds and preventing spurious content.

    \item We evaluate InfraDiffusion on two representative datasets (one masonry tunnel and two masonry bridges) featured by sparsity and geometric irregularity. By combining InfraDiffusion with the SAM for prompt-based zero-shot segmentation, we demonstrate substantial improvements in brick-level segmentation accuracy.
\end{enumerate}

The paper is organised as follows. Section 2 introduces the InfraDiffusion pipeline in detail, including the virtual camera-based depth map projection and the adaptation of DDNM in InfraDiffusion for zero-shot depth restoration. Section 3 describes the two datasets used in this study: a masonry tunnel and three masonry bridges, both of which present real-world challenges in point cloud sparsity and geometric irregularity. Section 4 evaluates the performance of InfraDiffusion and prompt-based segmentation for assessing restoration quality. Finally, Section 5 concludes the paper with a discussion of our findings and potential directions for future research.

\section{Methodology}\label{sec2}

\subsection{Virtual camera projection of point clouds}

\begin{figure*}[t!]
    \centering
    \includegraphics[width=\linewidth]{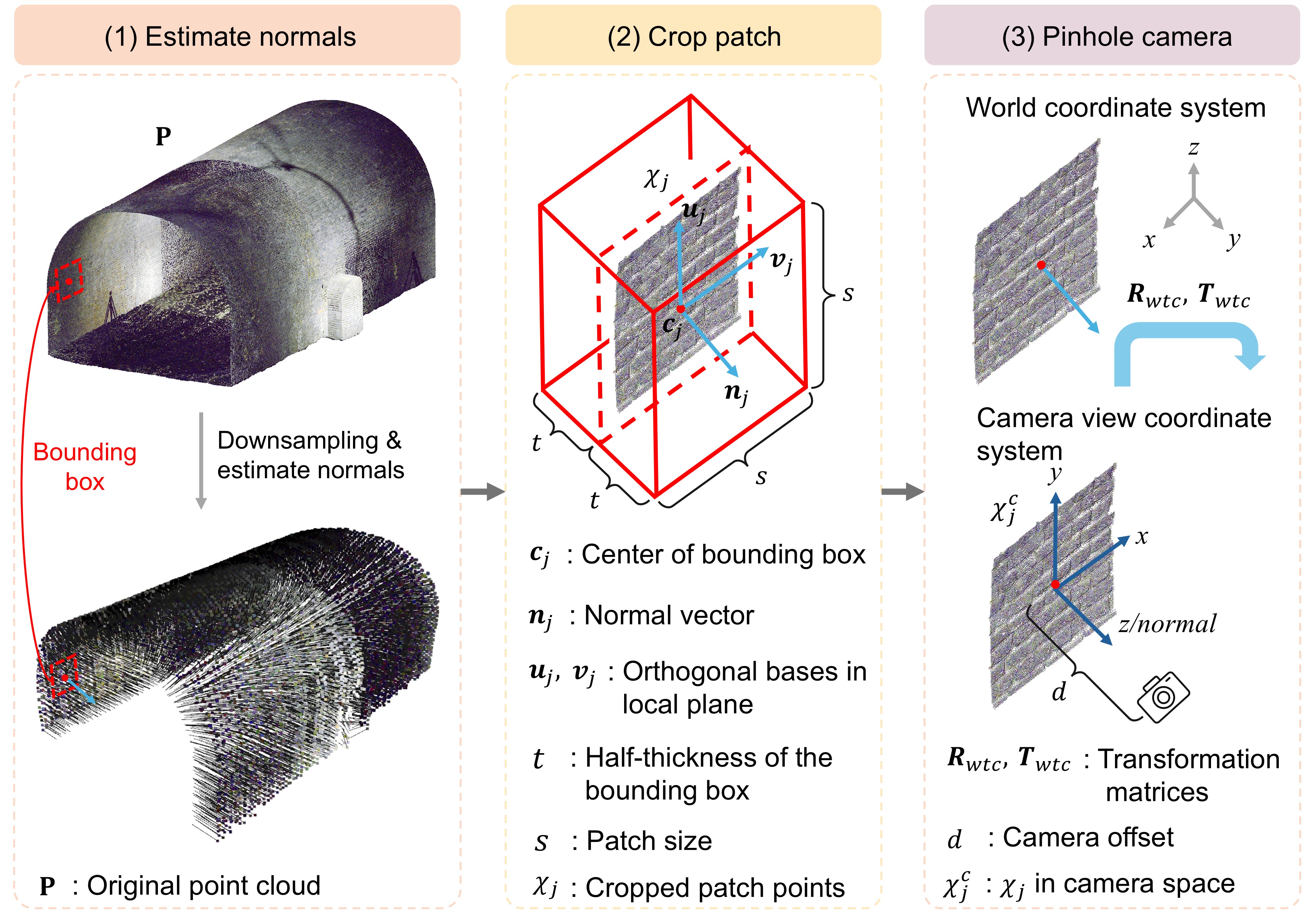}
    \caption{Overview of the patch extraction and projection using the pinhole camera model.}
    \label{fig: patch_projection}
\end{figure*}

As illustrated in Figure~\ref {fig: patch_projection}, the general framework which transfers infrastructure point clouds into depth maps is achieved in three steps: (1) normal estimation, (2) patch cropping, and (3) virtual camera projection (pinhole camera).

\paragraph{\textbf{Normal estimation}} To improve computational efficiency, voxel-based downsampling with a voxel size of $0.2$ m is first applied to the original point cloud $ \mathbf{P} = \{ \mathbf{x}_i \in \mathbb{R}^3 \}_{i=1}^{N}$ to obtain a sparse representation, where $N$ represents the total point number. Surface normals are then estimated on the downsampled point cloud using a ball query with a radius of $0.4$ m, implemented via the Open3D library \citep{zhou2018open3d}. This process generates a set of patch centre points $\mathbf{c}_j \in \mathbb{R}^3$ and their corresponding normal vectors $\mathbf{n}_j$, where $j$ represents the index of the centre points. These parameter values were selected based on empirical trials on real masonry infrastructure point clouds, which balance computational efficiency with robust normal estimation.

\paragraph{\textbf{Patch cropping}} For each $\mathbf{c}_j$, we define a bounding box centered at $\mathbf{c}_j$, as illustrated in the red box in Figure~\ref{fig: patch_projection}. Each patch has a square face with side length $s$ and thickness $2t$. The specific values of these parameters are listed in Table~\ref{tab:camera_params}. The added thickness accounts for geometric variation and ensures that sufficient points are captured near corners or regions where the surface is not shell-like. The cropped points $\mathcal{X}_j$ in the bounding box is defined as:

\begin{equation}\label{eq:bounding_box}
\begin{split}
\mathcal{X}_j = \left\{ \mathbf{x}_{ij} \in \mathbf{P} \,\middle|\,
    \left|(\mathbf{x}_{ij} - \mathbf{c}_j)^\top \mathbf{u}_j\right| \leq \frac{s}{2}, \;
    \left|(\mathbf{x}_{ij} - \mathbf{c}_j)^\top \mathbf{v}_j\right| \leq \frac{s}{2}, \right. \\
    \left. \left|(\mathbf{x}_{ij} - \mathbf{c}_j)^\top \mathbf{n}_j\right| \leq t \right\}
\end{split}
\end{equation}
where $\{\mathbf{u}_j, \mathbf{v}_j\}$ form an orthonormal basis spanning the local tangent plane orthogonal to $\mathbf{n}_j$ as shown in Figure~\ref {fig: patch_projection}. The parameter choices were determined through trial-and-error on real masonry infrastructure datasets to leverage between local completeness and boundary clarity of brick and mortar.

\begin{table}[t!]
\centering
\caption{Patch extraction and virtual camera parameters.}
\vspace{0.5em}
\begin{tabular}{lll}
\toprule
\textbf{Module} & \textbf{Parameter} & \textbf{Value} \\
\midrule
\multirow{2}{*}{Crop patch} 
    & \( s \) & 0.8 m \\
    & \( t \) & 0.25 m \\
\midrule
\multirow{4}{*}{Pinhole camera} 
    & \( d \) & 0.8 m \\
    & \( (f_x, f_y) \) & (400, 400) pixels \\
    & $(H, W)$ & \( (256 \times 256) \) pixels \\
    & \( (c_x, c_y) \) & (128, 128) pixels \\
\bottomrule
\end{tabular}
\label{tab:camera_params}
\end{table}

\paragraph{\textbf{Pinhole camera}} Each $\mathcal{X}_j$ is projected onto a sparse and noisy 2D depth map, i.e., $\mathbf{\tilde{y}}_j \in \mathbb{R}^{H \times W}$, using a virtual pinhole camera model. The virtual camera is positioned a fixed distance $d$ from $\mathbf{c}_j$, with its optical axis aligned with $\mathbf{n}_j$, as shown in Figure~\ref {fig: patch_projection}. The camera origin $\mathbf{o}_j$ can thus be computed as:

\begin{equation}
    \mathbf{o}_j = \mathbf{c}_j + d \cdot \mathbf{n}_j
\end{equation}

The camera view coordinate system is constructed using $\{\mathbf{n}_j, \mathbf{u}_j, \mathbf{v}_j\}$. This defines the rotation matrix \( \mathbf{R}_{wtc} \in \mathbb{R}^{3 \times 3} \) and the translation vector \( \mathbf{T}_{wtc} = \mathbf{o}_j \), such that the world-to-camera transformation is:

\begin{equation}
    \mathbf{x}^c_{ij} = \mathbf{R}_{wtc} (\mathbf{x}_{ij} - \mathbf{T}_{wtc})
\end{equation}
where $\mathbf{x}^c_{ij} = (x^c_{ij}, \, y^c_{ij}, \, z^c_{ij})$ is the transformed point in the camera view coordinate system to form local patch $\mathcal{X}^c_j$.

$\mathcal{X}^c_j$ is then projected onto the image plane using the pinhole camera model with focal lengths \( f_x, f_y \) and principal point \( (c_x, c_y) \):
\begin{equation}
    \begin{bmatrix}
        u \\
        v
    \end{bmatrix}
    =
    \begin{bmatrix}
        f_x \cdot \frac{x^c_{ij}}{z^c_{ij}} + c_x \\
        f_y \cdot \frac{y^c_{ij}}{z^c_{ij}} + c_y
    \end{bmatrix}
\end{equation}
where the depth value at pixel $(u, v)$ is taken as the $z$-coordinate of the point in the camera frame. The image resolution $(H, W)$, $d$, \( f_x, f_y \), and \( (c_x, c_y) \) are listed in Table~\ref{tab:camera_params}. These values are selected to maximise coverage of $\mathcal{X}^c_j$ whilst preserving the necessary details of individual bricks.

\begin{figure*}[t!]
    \centering
    \includegraphics[width=\linewidth]{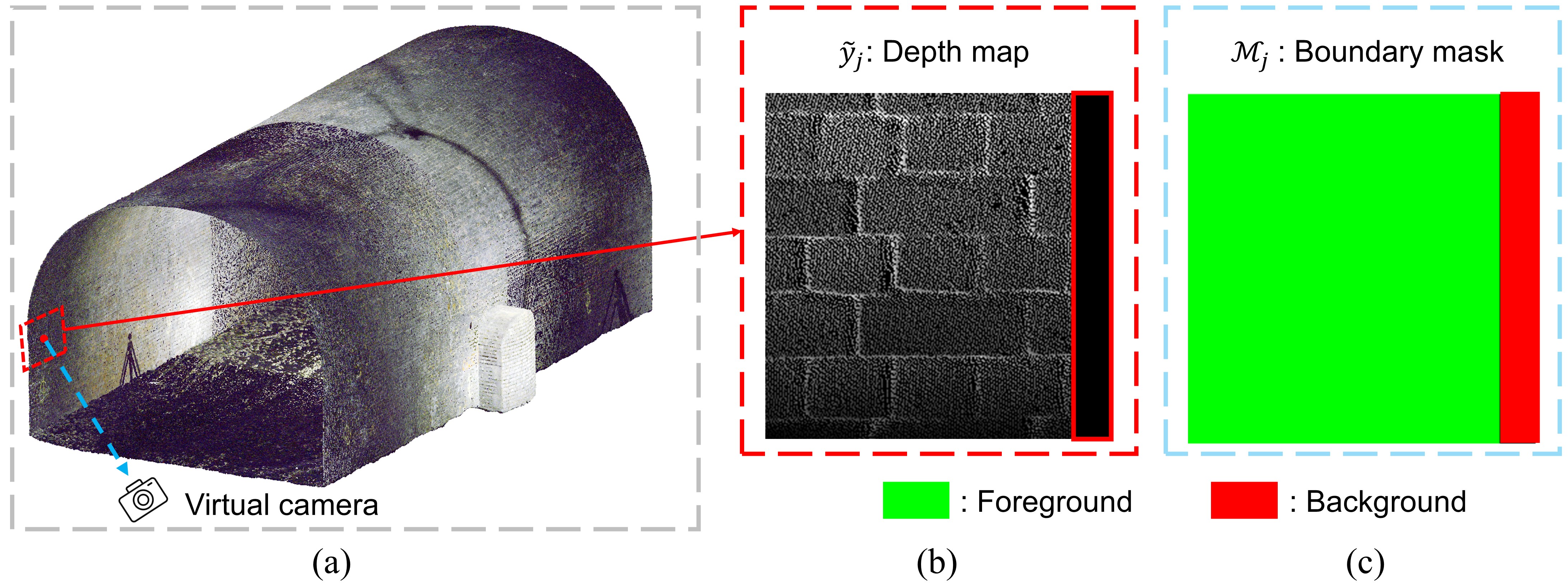}
    \caption{Boundary mask $\mathcal{M}_j$ extraction. (a) represents a virtual camera projection near the infrastructure boundary; (b) shows depth mask $\mathbf{\tilde{y}}_j$ with missing regions shown by the solid red box; (c) visualises the boundary mask $\mathcal{M}_j$ where the foreground and background are shown in green and red respectively.}
    \label{fig: boundary_mask}
\end{figure*}

Due to the fixed field of view of the virtual camera, the resulting image plane may include regions not covered by any 3D points from $\mathcal{X}^c_j$, particularly when the patch lies near the boundary of the structure. As illustrated in Figure~\ref{fig: boundary_mask}a, the projection produces $\mathbf{\tilde{y}}_j$ with partially empty regions, as shown in Figure~\ref{fig: boundary_mask}b (red solid box). These empty regions arise naturally from projecting irregular masonry geometries, and if left unaddressed, subsequent IR algorithms (e.g., DDNM) may introduce spurious bricks in empty regions or blur true structural boundaries.

To address this, we introduce a boundary mask $\mathcal{M}_j \subset \mathbb{R}^2$, defined as the axis-aligned 2D bounding box enclosing all projected pixels of $\mathcal{X}^c_j$:

\begin{equation}
    \mathcal{M}_j = \text{BBox}\left( \left\{ (u_i, v_i) \,\middle|\, \mathbf{x}^c_{ij} \in \mathcal{X}_j \right\} \right)
\end{equation}

As shown in Figure~\ref{fig: boundary_mask}c, pixels within $\mathcal{M}_j$ that correspond to valid projections are designated as the foreground (green), while pixels inside the mask but not covered by any projected point are labelled as the background (red). By explicitly separating valid and invalid regions, the boundary mask constrains restoration to physically meaningful areas, preventing the generation of non-existing bricks around structural edges.

\subsection{Zero-shot depth map restoration with DDNM}

To enable depth map restoration without task-specific training or clean GT supervision, we adapt the DDNM \citep{wang2022zero} conditioned on the boundary mask $\mathcal{M}_j$. DDNM is a zero-shot IR framework that leverages the generative capacity of pre-trained DDPMs. DDPMs can learn statistical dependencies among pixels from large image datasets, thereby capturing strong contextual priors that can be exploited to reconstruct missing or degraded regions in depth maps.

This section begins with a brief overview of the classical IR problem. We then introduce the range–null space decomposition framework. Following this, we review the theoretical foundations of DDPM \citep{ho2020denoising} and DDIM \citep{song2020denoising}, where DDIM eliminates the Markov assumption to improve sampling efficiency. Finally, we describe the DDNM algorithm and detail how it is adapted herein to restore sparse and noisy depth maps projected from masonry infrastructure point clouds.

\subsubsection{Background of IR problem}

IR is a long-standing inverse problem involving the recovery of a clean image $\mathbf{y}_j \in \mathbb{R}^{H \times W}$ from a degraded observation \citep{richardson1972bayesian, 6323527} (denoted as $\tilde{\mathbf{y}}_j$ in this paper). The degradation is typically modelled by a known linear operator $\mathbf{A}_j$, which represents corruption such as downsampling, missing pixels, blur, additive noise, or combinations thereof. The general observation model is expressed as:

\begin{equation}\label{eq: IR_problem}
    \tilde{\mathbf{y}}_j = \mathbf{A}_j \mathbf{y}_j + \mathbf{n}
\end{equation}
where $\mathbf{n} \in \mathbb{R}^{H \times W} \sim \mathcal{N} (\textbf{0}, \sigma_y^2 \mathbf{I})$ denotes additive noise. Here, we address two canonical subproblems of IR, namely inpainting (data sparsity) and noise, by restoring $\mathbf{y}_j$ through InfraDiffusion. The degradation operator $\mathbf{A}_j$ is instantiated as a binary spatial mask derived from the projection of the point cloud, indicating which pixels in $\tilde{\mathbf{y}}_j$ correspond to valid 3D points and which are unobserved.

Restoring $\mathbf{y}_j$ from $\tilde{\mathbf{y}}_j$ is inherently ill-posed due to severe sparsity and noise during projection and point cloud data collection. Traditional non-generative IR approaches \citep{dong2014learning} often rely on task-specific training with paired data (e.g., $\mathbf{y}_j$ and $\tilde{\mathbf{y}}_j$), which is impractical for infrastructure due to the dearth of $\mathbf{y}_j$, and often yields poor results in real-world scenarios. To overcome this, we adapt the zero-shot DDNM restoration method guided by the strong generative priors of DDPM, enabling robust and accurate restoration without supervised training.

\subsubsection{Range-null space decomposition in noise-free IR problem}

We begin our analysis with the noise-free IR setting, where the simplification allows for a more interpretable formulation of the range-null space decomposition, which is represented as:

\begin{equation}
\tilde{\mathbf{y}}_j = \mathbf{A}_j \mathbf{y}_j
\end{equation}

The objective is to reconstruct a plausible depth image $\hat{\mathbf{y}}_j$ that satisfies two essential conditions:
\begin{align}
\textit{Consistency:} \quad & \mathbf{A}_j \hat{\mathbf{y}}_j = \tilde{\mathbf{y}}_j \\
\textit{Realness:} \quad & \hat{\mathbf{y}}_j \sim q(\mathbf{y})
\end{align}
where \textit{Consistency} ensures $\hat{\mathbf{y}}_j$ satisfies the degradation relationship, and \textit{Realness} determines whether $\hat{\mathbf{y}}_j$ is sampled from the clean depth image distribution, i.e., $ q(\mathbf{y})$.

To analyse the solution space, DDNM \citep{wang2022zero} decomposes $\hat{\mathbf{y}}_j$ into its range and null spaces of $\mathbf{A}_j$:
\begin{equation}\label{eq: range-null dp}
\hat{\mathbf{y}}_j = \mathbf{A}^{\dagger}_j \mathbf{A}_j \hat{\mathbf{y}}_j + (\mathbf{I} - \mathbf{A}_j^{\dagger} \mathbf{A}_j) \hat{\mathbf{y}}_j
\end{equation}
where $\mathbf{A}_j^{\dagger}$ is the Moore–Penrose pseudoinverse of $\mathbf{A}_j$, which can be derived via singular value decomposition \citep{golub1971singular}. $\mathbf{A}_j^{\dagger}$ for the IR problem has been properly introduced in DDPM \citep{wang2022zero}, which is omitted in this paper for simplicity. By letting:
\begin{equation}
\mathbf{y}_{j}^{\text{range}} := \mathbf{A}_j^{\dagger} \mathbf{A}_j \hat{\mathbf{y}}_j, \quad
\mathbf{y}_{j}^{\text{null}} := (\mathbf{I} - \mathbf{A}_j^{\dagger} \mathbf{A}_j) \bar{\mathbf{y}}_j
\end{equation}
where the $\hat{\mathbf{y}}_j$ in the second item of Equation~\ref{eq: range-null dp} is changed to $\bar{\mathbf{y}}_j$ as it is generated from DDPM. The Equation~\ref{eq: range-null dp} is re-written as:
\begin{equation}
\hat{\mathbf{y}}_j = \mathbf{y}_{j}^{\text{range}} + \mathbf{y}_{j}^{\text{null}}
\end{equation}

Applying the degradation operator $\mathbf{A}_j$ to both sides, we obtain:
\begin{equation}
\mathbf{A}_j \hat{\mathbf{y}}_j = \mathbf{A}_j \mathbf{y}_{j}^{\text{range}} + \mathbf{A}_j \mathbf{y}_{j}^{\text{null}} = \mathbf{A}_j \mathbf{y}_{j}^{\text{range}}= \mathbf{A}_j \hat{\mathbf{y}}_j
\end{equation}
since $\mathbf{A}_j \mathbf{y}_{j}^{\text{null}} = \mathbf{A}_j (\mathbf{I} - \mathbf{A}_j^{\dagger} \mathbf{A}_j) \bar{\mathbf{y}}_j = \mathbf{0}$, $\hat{\mathbf{y}}_i$ is not influenced by any $\bar{\mathbf{y}}_j$ after being applied by $\mathbf{A}_j$; all observable content is confined to $\mathbf{y}_{j}^{\text{range}}$, which guarantees the \textit{Consistency} constraint. However, $\mathbf{y}_{j}^{\text{null}}$ determines whether it is perceptually plausible, i.e., whether it satisfies the \textit{Realness} criterion by resembling a sample from $\hat{\mathbf{y}}_j \sim q(\mathbf{y})$. To achieve \textit{Realness}, DDNM leverages the generative capability of DDPM to sample $\bar{\mathbf{y}}_j$.

\subsubsection{Review of DDPM and DDIM}

DDPMs \citep{ho2020denoising} are a class of generative models that reconstruct data by reversing a gradual noising process. The coral idea is to learn how to transform random noise into realistic image samples by exploiting the statistical dependencies observed in training data. Let $\mathbf{y}_0 \in \mathbb{R}^{H \times W}$ denote the original clean image. The forward diffusion process progressively adds Gaussian noise to $\mathbf{y}_0$ over $T$ steps, eventually converting it into approximately pure noise drawn from a standard Gaussian distribution: 

\begin{equation}
    q(\mathbf{y}_t \mid \mathbf{y}_{t-1}) = \mathcal{N}(\mathbf{y}_t; \sqrt{1 - \beta_t} \, \mathbf{y}_{t-1}, \, \beta_t \mathbf{I})
\end{equation}
where $\{ \beta_t \}_{t=1}^T$ is a predefined variance schedule. By using the reparameterization trick, noisy samples at step $t$ can be written as:

\begin{equation}
    \mathbf{y}_t = \sqrt{\bar{\alpha}_t} \, \mathbf{y}_0 + \sqrt{1 - \bar{\alpha}_t} \, \boldsymbol{\epsilon}, \quad \boldsymbol{\epsilon} \sim \mathcal{N}(\mathbf{0}, \mathbf{I})
\end{equation}
where $\bar{\alpha}_t = \prod_{s=1}^t (1 - \beta_s)$. The generative process aims to reverse the diffusion process by sampling $\mathbf{y}_{t-1}$ from $\mathbf{y}_{t}$ via Bayes' rule:
\begin{align}
    p(\mathbf{y}_{t-1} \mid \mathbf{y}_t, \mathbf{y}_0) &= q(\mathbf{y}_t \mid \mathbf{y}_{t-1}, \mathbf{y}_0) \cdot \frac{q(\mathbf{y}_{t-1} \mid \mathbf{y}_0)}{q(\mathbf{y}_t \mid \mathbf{y}_0)}\\
    &= \mathcal{N}(\mathbf{y}_{t-1}; \boldsymbol{\mu}_t(\mathbf{y}_t, \mathbf{y}_0), \sigma_t^2 \mathbf{I})
\end{align}
where $p(\mathbf{y}_{t-1} \mid \mathbf{y}_t, \mathbf{y}_0)$ still represents a Gaussian distribution. The mean $\boldsymbol{\mu}_t(\mathbf{y}_t, \mathbf{y}_0)$ and variance $\sigma_t^2$ are derived from the forward process and can be analytically computed given $\{ \beta_t \}_{t=1}^T$. In practice, rather than learning the full reverse distribution directly, DDPM trains a neural network $\boldsymbol{\epsilon}_\theta(\mathbf{y}_t, t)$ to predict the added noise $\boldsymbol{\epsilon}$. The training objective then becomes:

\begin{equation}\label{eq: loss_DDPM}
    \mathcal{L}_{DDPM} = \mathbb{E}_{\mathbf{y}_0, \boldsymbol{\epsilon}, t} \left[ \left\| \boldsymbol{\epsilon} - \boldsymbol{\epsilon}_\theta(\sqrt{\bar{\alpha}_t} \, \mathbf{y}_0 + \sqrt{1 - \bar{\alpha}_t} \, \boldsymbol{\epsilon}, \, t) \right\|^2 \right]
\end{equation}

Equation~\ref{eq: loss_DDPM} enables the network to learn to remove noise at arbitrary timesteps $t$, and forms the foundation for image generation and restoration via iterative denoising.
 
While DDPM achieves high-quality generative performance, its sampling process is slow due to the need for thousands of denoising steps. This inefficiency arises from its reliance on a Markovian reverse process, i.e., $q(\mathbf{y}_t \mid \mathbf{y}_{t-1}, \mathbf{y}_0) = q(\mathbf{y}_t \mid \mathbf{y}_{t-1})$, which requires step-wise resampling from Gaussian distributions.

To accelerate reverse sampling, DDIM \citep{song2020denoising} removes the Markov assumption and reformulates the reverse process as a non-stochastic transformation. This allows the use of an arbitrary set of decreasing timesteps $\{ \tau_1, \dots, \tau_K \} \subset \{1, \dots, T\}$, enabling larger step sizes in sampling. The DDIM sampling rule is given by:

\begin{align}
    \mathbf{y}_{\tau_{k-1}} = & \; \sqrt{\bar{\alpha}_{\tau_{k-1}}} \left( 
        \frac{\mathbf{y}_{\tau_k} - \sqrt{1 - \bar{\alpha}_{\tau_k}} \cdot \boldsymbol{\epsilon}_\theta(\mathbf{y}_{\tau_k}, \tau_k)}
             {\sqrt{\bar{\alpha}_{\tau_k}}} \right) \notag \\
    & + \sqrt{1 - \bar{\alpha}_{\tau_{k-1}} - \sigma_{\tau_k}^2} \cdot \boldsymbol{\epsilon}_\theta(\mathbf{y}_{\tau_k}, \tau_k)
    + \sigma_{\tau_k} \cdot \boldsymbol{\epsilon}
\end{align}
where the setting of variance $\sigma_{\tau_k} \in [0, 1]$ controls the amount of stochasticity. DDIM preserves the forward process and training objective (gradient of $\mathcal{L}_{DDPM}$) of DDPM but replaces the stochastic differential equation (SDE) with an equivalent ordinary differential equation (ODE), enabling faster and controllable sampling.

\subsubsection{InfraDiffusion with boundary-constrained DDNM in noise IR problem}

We now consider the IR problem in the presence of noise, which is given by Equation~\ref{eq: IR_problem}. During the diffusion process, we denote $\mathbf{y}_{0|t}$ as the predicted clean image at timestep $t$ from DDPM. Following the DDNM framework \citep{wang2022zero}, the range–null space decomposition can be expressed as:  

\begin{equation}
\hat{\mathbf{y}}_{j,\,0|t} = \mathbf{A}_j^\dagger \tilde{\mathbf{y}}_j + (\mathbf{I} - \mathbf{A}_j^\dagger \mathbf{A}_j)\mathbf{y}_{0|t}
= \mathbf{y}_{0|t} - \mathbf{A}_j^\dagger (\mathbf{A}_j \mathbf{y}_{0|t} - \mathbf{A}_j \mathbf{y}_j) + \textcolor{red}{\mathbf{A}_j^\dagger \mathbf{n}}
\end{equation}
where the second term, i.e., $\mathbf{A}_j^\dagger (\mathbf{A}_j \mathbf{y}_{0|t} - \mathbf{A}_j \mathbf{y}_j)$, is a correction in the range-space that ensures \textit{Consistency}. $\textcolor{red}{\mathbf{A}_j^\dagger \mathbf{n}} \in \mathbb{R}^{H \times W}$ propagates the observation noise into $\hat{\mathbf{y}}_{j,\,0|t}$ and subsequently into $\mathbf{y}_{t-1}$ during sampling.

To align with the noise scale required by DDPM~\citep{ho2020denoising}, we adopt the DDNM formulation and reformulate the update as:
\begin{equation}
\hat{\mathbf{y}}_{j, \, 0|t} = \mathbf{y}_{0|t} - \boldsymbol{\Sigma}_t \mathbf{A}_j^\dagger ( \mathbf{A}_j \mathbf{y}_{0|t} - \tilde{\mathbf{y}}_j )
\end{equation}
where $\boldsymbol{\Sigma}_t$ is a time-dependent scaling matrix determined by $\sigma_y$ and the DDPM variance schedule (see \citep{wang2022zero} for full derivation).

Different from the original DDNM, InfraDiffusion incorporate $\textcolor{blue}{\mathbf{M}_j}$ as an extra condition to constrain the correction region. This restricts the correction region to valid projections, thereby preventing spurious generations in empty areas and ensuring that denoising and inpainting occur only within physically meaningful regions:
\begin{equation}
\hat{\mathbf{y}}_{j, \, 0|t} = \textcolor{blue}{\mathbf{M}_j} \odot \left( \mathbf{y}_{0|t} - \boldsymbol{\Sigma}_t \mathbf{A}_j^\dagger ( \mathbf{A}_j \mathbf{y}_{0|t} - \tilde{\mathbf{y}}_j ) \right) 
\end{equation}

During the sampling, $\hat{\mathbf{y}}_{j,0|t}$ is used to produce the $\mathbf{y}_{t-1}$, where any hallucinated content outside the valid support at step $t$ is propagated forward by both the noise prediction network and the range-space correction without $\textcolor{blue}{\mathbf{M}_j}$. Pixels outside $\mathcal{M}_j$ lie in the null space of $\mathbf{A}_j$ and, without constraints, are filled by the generative prior at every step. Applying $\mathcal{M}_j$ \emph{within} the update suppresses these contributions at each iteration, preserves sharp boundaries, and prevents the accumulation of boundary artefacts that a post-hoc mask cannot remove.

To perform efficient sampling, we adopt the DDIM framework~\citep{song2020denoising} instead of DDPM. The complete InfraDiffusion sampling procedure is summarised and explained in Algorithm~\ref{alg:ddnm_masked_ddim}.

\begin{algorithm}[t]
\caption{InfraDiffusion sampling with boundary masks}
\label{alg:ddnm_masked_ddim}
\begin{algorithmic}[1]
\State \textbf{Input:} Noisy image $\mathbf{y}_T \sim \mathcal{N}(\mathbf{0}, \mathbf{I})$, known data $\tilde{\mathbf{y}}_j$, degradation matrix $\mathbf{A}_j$, mask $\mathcal{M}_j$, step schedule $\{\tau_k\}_{k=1}^K$
\For{$k = K, \dots, 1$}
    \State $\mathbf{y}_{0|\tau_k} = \frac{1}{\sqrt{\bar{\alpha}_{\tau_k}}} \left( \mathbf{y}_{\tau_k} - \sqrt{1 - \bar{\alpha}_{\tau_k}} \cdot \boldsymbol{\epsilon}_\theta(\mathbf{y}_{\tau_k}, \tau_k) \right)$
    \State $\hat{\mathbf{y}}_{j,\,0|\tau_k} = \textcolor{blue}{\mathbf{M}_j} \odot \left( \mathbf{y}_{0|\tau_k} - \boldsymbol{\Sigma}_{\tau_k} \mathbf{A}_j^\dagger ( \mathbf{A}_j \mathbf{y}_{0|\tau_k} - \tilde{\mathbf{y}}_j ) \right)$
    \State $\begin{aligned}[t]
            \mathbf{y}_{\tau_{k-1}} =\; & \sqrt{\bar{\alpha}_{\tau_{k-1}}} \cdot \hat{\mathbf{y}}_{j,\,0|\tau_k} \\
            & + \sqrt{1 - \bar{\alpha}_{\tau_{k-1}} - \sigma_{\tau_k}^2} \cdot \boldsymbol{\epsilon}_\theta(\mathbf{y}_{\tau_k}, \tau_k) \\
            & + \sigma_{\tau_k} \cdot \boldsymbol{\epsilon}
        \end{aligned}$
    \Comment{$\boldsymbol{\epsilon} \sim \mathcal{N}(0, \mathbf{I})$}
\EndFor
\State \textbf{Return:} Reconstructed image $\mathbf{y}_0$
\end{algorithmic}
\end{algorithm}

\section{Dataset}

We evaluate the proposed InfraDiffusion framework on two types of masonry infrastructure: a historic railway tunnel and two masonry bridges. All point clouds were captured via terrestrial laser scanning (TLS) and semantically segmented into structural components. To avoid extensive computation while demonstrating our method, we sample a subset of representative chunks from the tunnel and randomly select patches across components in all datasets.

\begin{figure*}[t!]
    \centering
    \includegraphics[width=\linewidth]{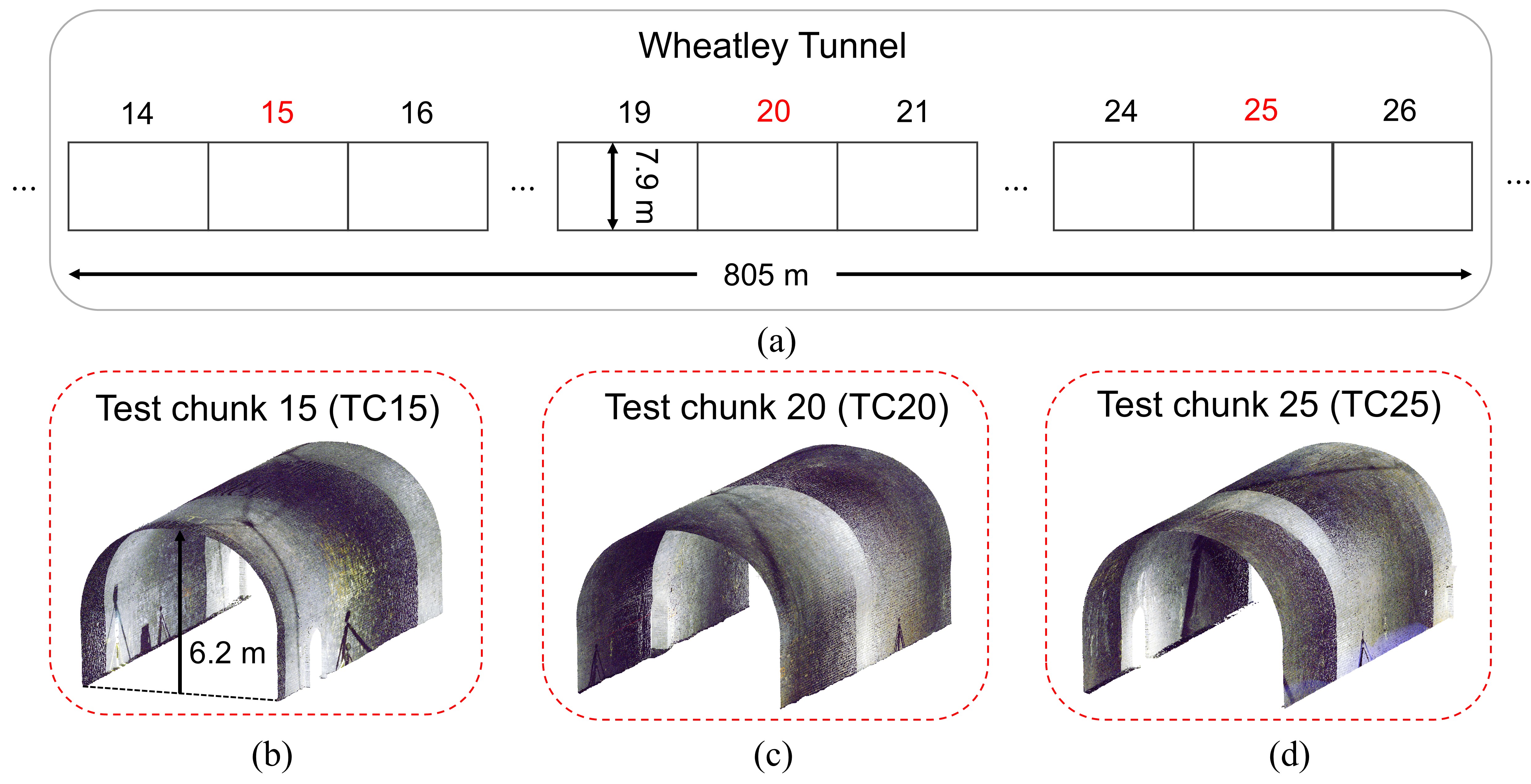}
    \caption{Wheatley Tunnel overview and selected test chunks. (a) Schematic plan view of the 805 m tunnel divided into 51 chunks, where chunks 15, 20, and 25 are selected for testing. (b)–(d) represent point cloud views of the selected chunks (e.g., TC15, TC20, and TC25, respectively).}
    \label{fig: wheatley_overview}
\end{figure*}

\subsection{Wheatley Tunnel}\label{sec: wheatley_dataset}

The Wheatley Tunnel is a historic masonry railway tunnel located on the Halifax High Level line, spanning a total length of 805 m. The original point cloud was collected by National Highways as part of a structural health monitoring initiative. According to inspection records, the tunnel remains in overall fair condition but exhibits widespread dampness, soft mortar loss, and calcite deposits throughout the bore lining. Signs of spalling are also observed in the arch and walls.

As shown in Figure~\ref{fig: wheatley_overview}a-d, the tunnel was divided into 51 contiguous chunks, each approximately 15 m in length, due to the substantial volume of point cloud data. For evaluation, we selected three representative chunks (e.g., TC15, TC20, and TC25), covering structurally diverse regions of the tunnel. The typical cross-sectional dimensions of the tunnel are approximately 7.9 m in width and 6.2 m in height (Figure~\ref{fig: wheatley_overview}a and b).

\begin{table}[t]
    \centering
    \caption{Summary of selected Wheatley Tunnel chunks with point counts, structural composition, and number of extracted patches.}
    \label{tab: wheatley_chunk_summary}
    \begin{tabular}{lcccc}
        \toprule
        \textbf{Chunk} & \textbf{Points} & \textbf{Arch} & \textbf{Wall} & \textbf{Patch} \\
        \midrule
        TC15 & 55,821,396 & 1 & 2 & 36 \\
        TC20 & 32,278,956 & 1 & 2 & 25 \\
        TC25 & 37,756,228 & 1 & 2 & 39 \\
        \bottomrule
    \end{tabular}
\end{table}

\begin{figure*}[ht!]
    \centering
    \includegraphics[width=0.75\linewidth]{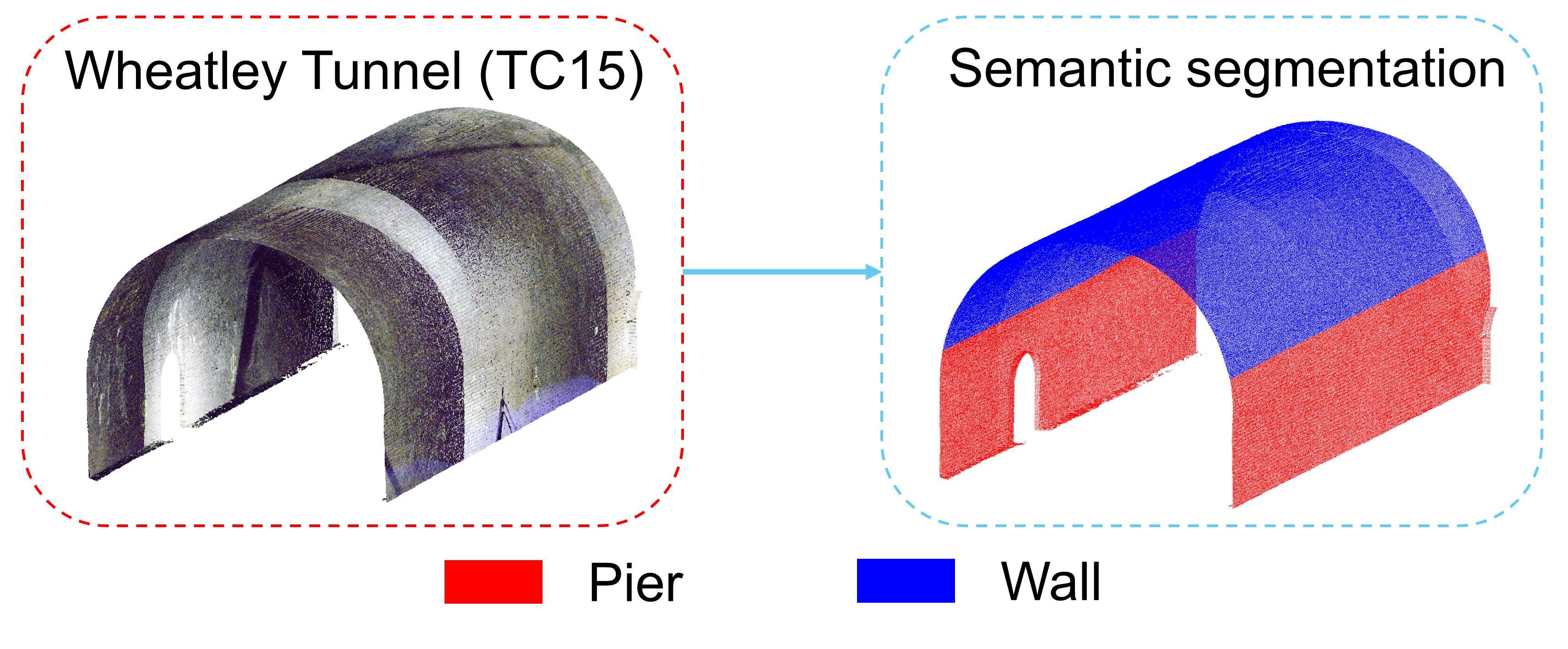}
    \caption{Semantic segmentation of the Wheatley Tunnel (TC15) point cloud. The arch and two walls are labelled in blue and red, respectively.}
    \label{fig: wheatley_seg}
\end{figure*}

\begin{figure*}[ht!]
    \centering
    \includegraphics[width=\linewidth]{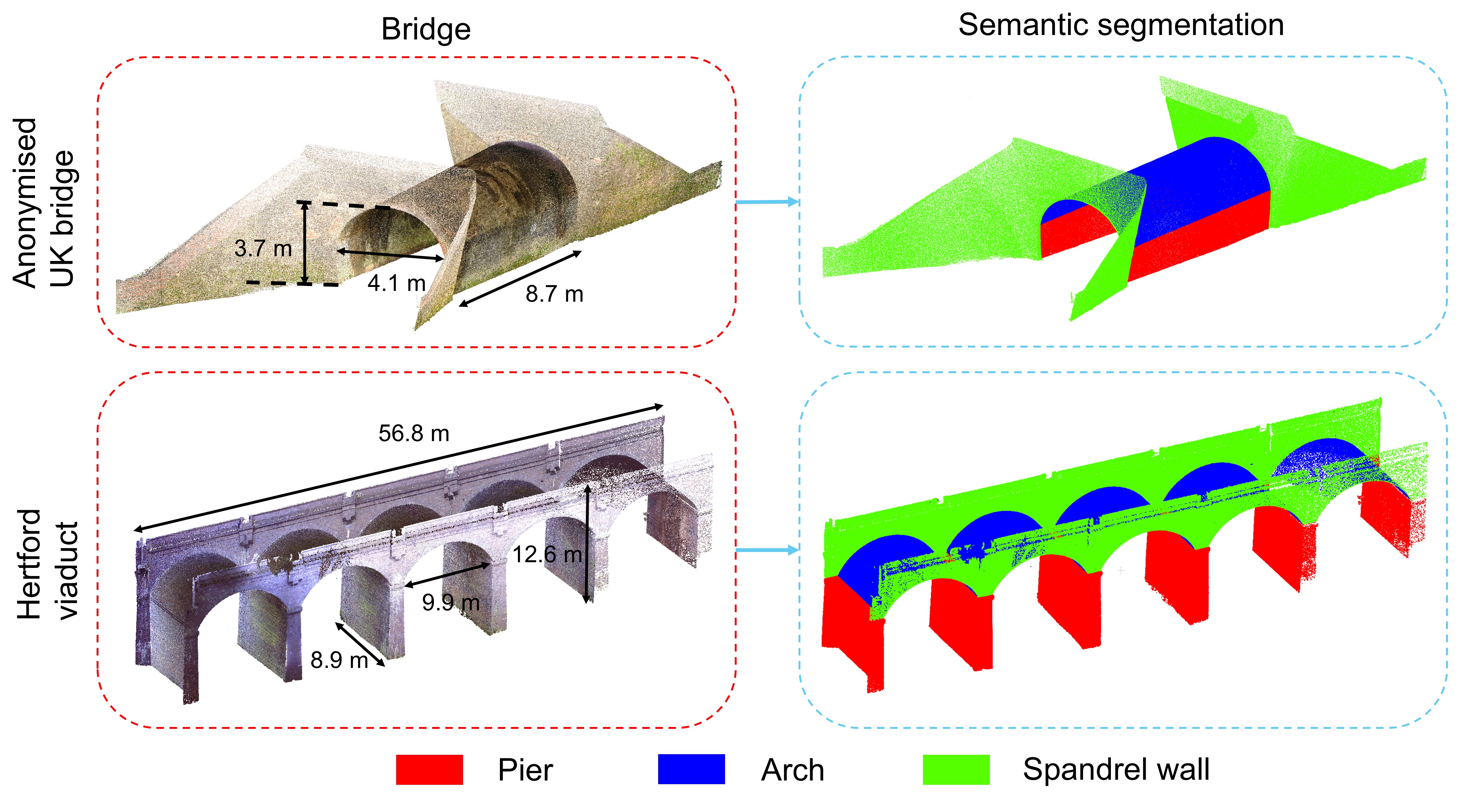}
    \caption{Point clouds and semantic segmentation of anonymised UK bridge and Hertford Viaduct. Dimensions annotated in the raw point cloud aid understanding of the structural scale and data volume. The arch, pier, and spandrel wall are coloured red, blue, and green, respectively.}
    \label{fig: bridge_segmentation}
\end{figure*}

To ensure unbiased coverage across different structural components, $\mathbf{c}_j$ (e.g., patch centre as shown in Figure~\ref{fig: patch_projection}) is randomly sampled from the full tunnel point cloud, without preference toward any specific region. Due to the large number of available patches, we sample a total of 100 patches from all three chunks for evaluation, given that they belong to the same tunnel. The distribution of selected patches across the three chunks is shown in Table~\ref{tab: wheatley_chunk_summary}. Each selected chunk includes a typical masonry composition of one arch barrel and two supporting walls, which are semantically segmented into distinct regions for analytical purposes. As shown in Figure~\ref{fig: wheatley_seg}, the segmentation allows us to assess how projection quality and IR performance vary across these structural elements. Such differences are often introduced by the varying distance between the LiDAR scanner and the scanned surface; regions farther from the scanner (e.g., crown or upper arch) tend to produce noisier and more diffuse depth images than closer surfaces (e.g., walls).

\subsection{Masonry bridges}

We also evaluate our method on two representative masonry bridge point clouds (Figure~\ref{fig: bridge_segmentation}): (1) an anonymised UK single-span masonry bridge acquired from an external industry partner, and (2) Hertford Viaduct, which is a multi-span masonry bridge \citep{jing2022segmentation}. Both bridges were segmented into structural components (arches, piers, and spandrel walls) to compare the depth image quality in different components. Figure~\ref{fig: bridge_segmentation} also shows the overall geometry and scale of each bridge, which includes key structural dimensions. The semantic segmentation uses blue for arches, red for piers, and green for spandrel walls to distinguish structural components across the bridge geometry.

\begin{table}[t]
    \centering
    \caption{Summary of masonry bridge datasets with point counts, component counts, and number of patches sampled.}
    \label{tab: bridge_summary}
    \begin{tabular}{lccccc}
        \toprule
        \textbf{Bridge} & \textbf{Points} & \textbf{Arch} & \textbf{Pier} & \textbf{Spandrel wall} & \textbf{Patch} \\
        \midrule
        \makecell[l]{Anonymised\\UK bridge} & 74,183,011 & 1 & 2 & 2 & 60 \\
        \makecell[l]{Hertford\\Viaduct}     & 98,629,976 & 5 & 6 & 2 & 140 \\
        \bottomrule
    \end{tabular}
\end{table}

Table~\ref{tab: bridge_summary} summarises the number of points, structural components, and sampled patches for each bridge. Patch selection was randomly conducted across all components to ensure structural diversity while maintaining a tractable evaluation size.

\section{Experiments and results}\label{sec4}

\subsection{Experiment Settings}

We employ a pre-trained, unconditional DM released by OpenAI \citep{dhariwal2021diffusion} as the generative prior for InfraDiffusion. All experiments are conducted in a zero-shot manner, without any task-specific fine-tuning or additional training.

Input depth images with a fixed resolution of $256 \times 256$ (see Table~\ref{tab:camera_params}) are normalised to the $[0, 1]$ range, which was empirically found to yield better restoration quality than normalising to $[-1, 1]$. The measurement noise in $\mathbf{y}$ is modelled using a standard deviation of $\sigma_y = 0.16$, which provides a stable assumption for subsequent restoration. Inference is performed on a laptop equipped with an NVIDIA RTX 4090 GPU (16 GB VRAM).

The model architecture consists of a U-Net backbone with a channel depth of 256, two residual blocks per level, and self-attention mechanisms applied at spatial resolutions of 32, 16, and 8, where more details can be found at \cite{dhariwal2021diffusion}. The diffusion process utilizes a linear $\beta$ schedule ranging from $\beta_\text{start} = 10^{-4}$ to $\beta_\text{end} = 0.02$ over 1000 steps. For sampling, we adopt the DDIM inversion strategy, reducing the number of denoising steps to 100 using 10-step intervals to accelerate inference.

\begin{figure*}[t!]
    \centering
    \includegraphics[width=\linewidth]{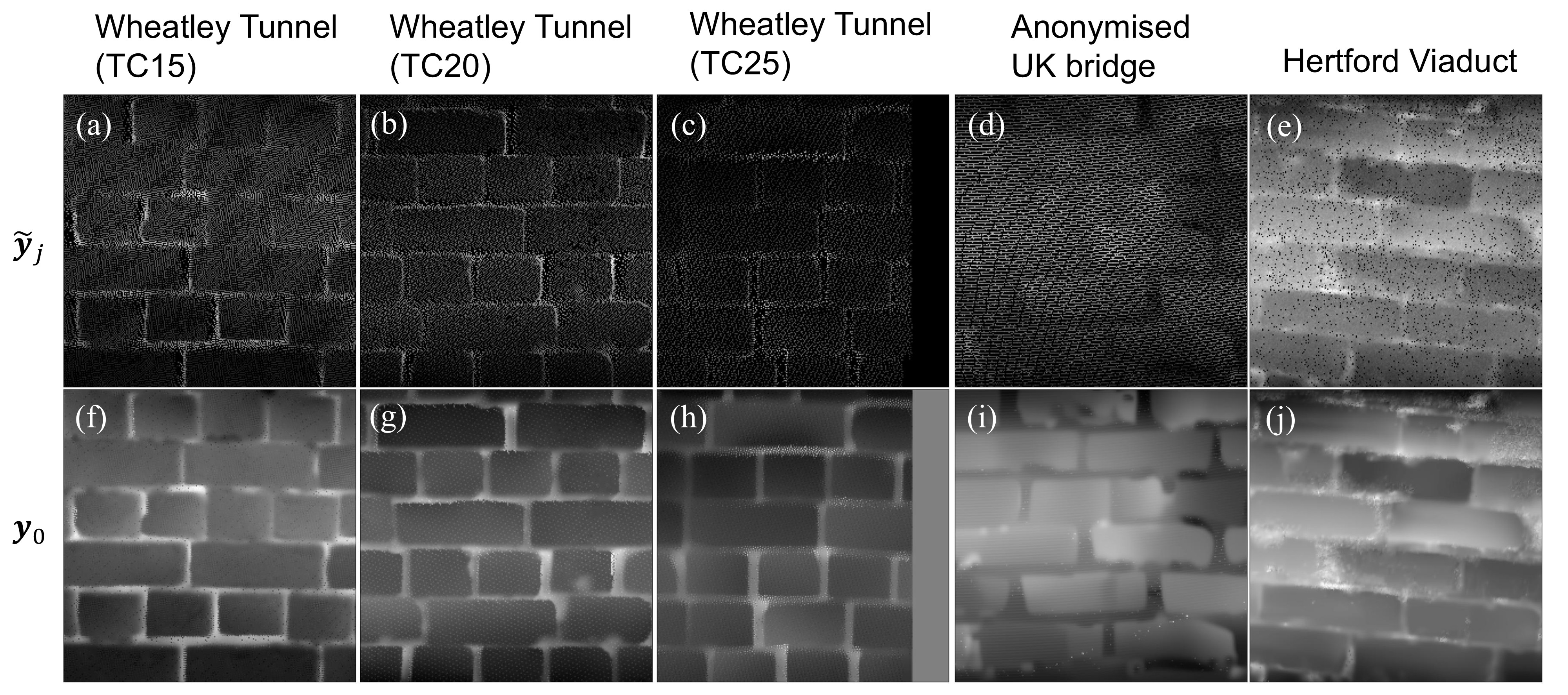}
    \caption{IR results for arch components across five infrastructures with $\sigma_y = 0.16$. (a)–(e) show degraded inputs $\tilde{\mathbf{y}}_j$, and (f)–(j) present restored images $\mathbf{y}_0$. (a, f), (b, g), and (c, h) are randomly sampled from Wheatley Tunnel (TC15, TC20, and TC25). (d, i) and (e, j) correspond to anonymised UK bridge and Hertford Viaduct.}
    \label{fig: inpainting_arch}
\end{figure*}

\subsection{Depth map restoration results}

\begin{figure*}[t!]
    \centering
    \includegraphics[width=\linewidth]{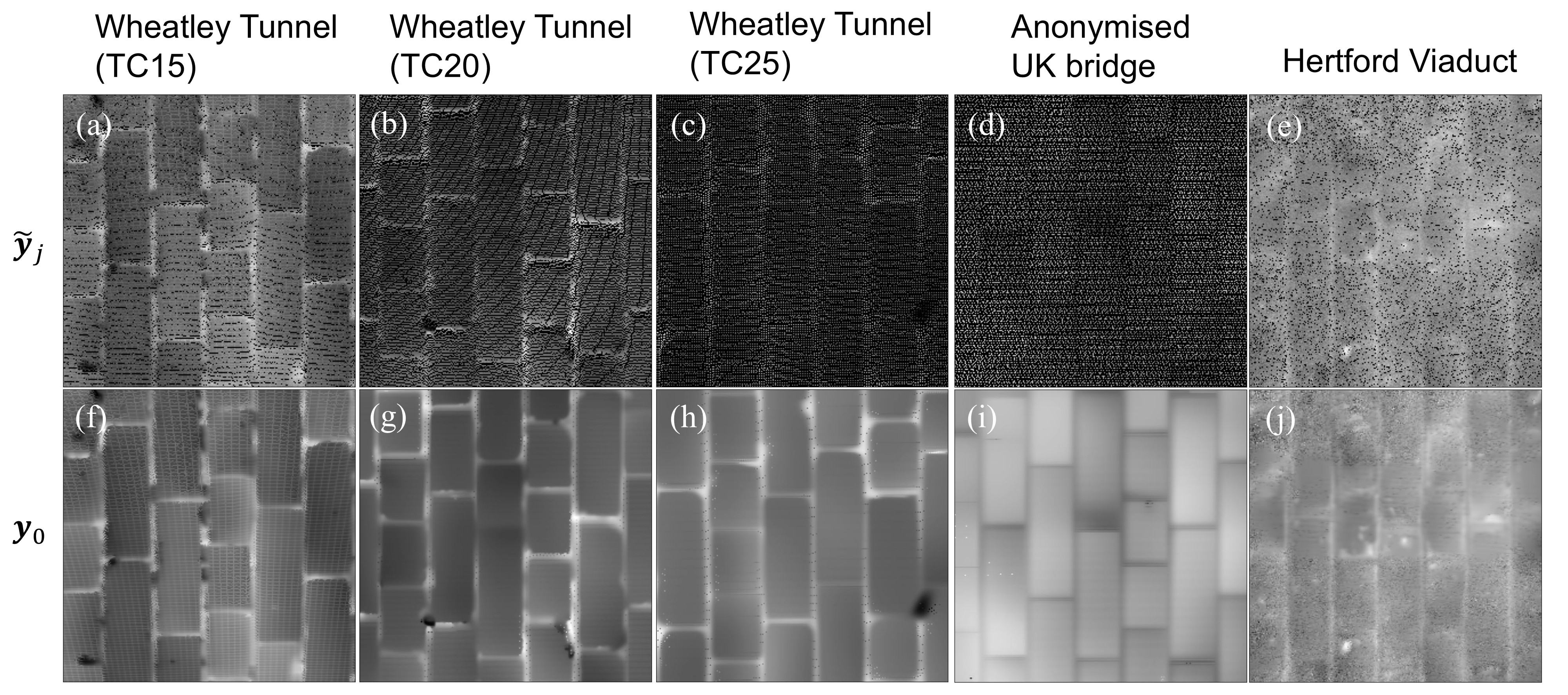}
    \caption{IR results for wall/pier components across five infrastructures with $\sigma_y = 0.16$. (a)–(e) show degraded inputs $\tilde{\mathbf{y}}_j$, and (f)–(j) present restored images $\mathbf{y}_0$. (a, f), (b, g), and (c, h) are randomly sampled from Wheatley Tunnel (TC15, TC20, and TC25). (d, i) and (e, j) correspond to anonymised UK bridge and Hertford Viaduct.}
    \label{fig: inpainting_pier}
\end{figure*}

\begin{table}[t!]
    \centering
    \caption{Patch indices selected for qualitative IR evaluation across arch and wall/pier components. Arch patches are visualised in Figure~\ref{fig: inpainting_arch}a-j and wall/pier patches in Figure~\ref{fig: inpainting_pier}a-j.}
    \resizebox{\textwidth}{!}{%
    \label{tab:patch_indices}
    \begin{tabular}{lcc}
        \toprule
        \textbf{Infrastructure} & \textbf{Arch patch indices} & \textbf{Wall/Pier patch indices} \\
        \midrule
        Wheatley Tunnel (TC15) & 1828 & 969 \\
        Wheatley Tunnel (TC20) & 807  & 270 \\
        Wheatley Tunnel (TC25) & 2352 & 986 \\
        Anonymised UK bridge & 509  & 269 \\
        Hertford Viaduct & 703  & 145 \\
        \bottomrule
    \end{tabular}
    }
\end{table}

\begin{figure*}[t!]
    \centering
    \includegraphics[width=\linewidth]{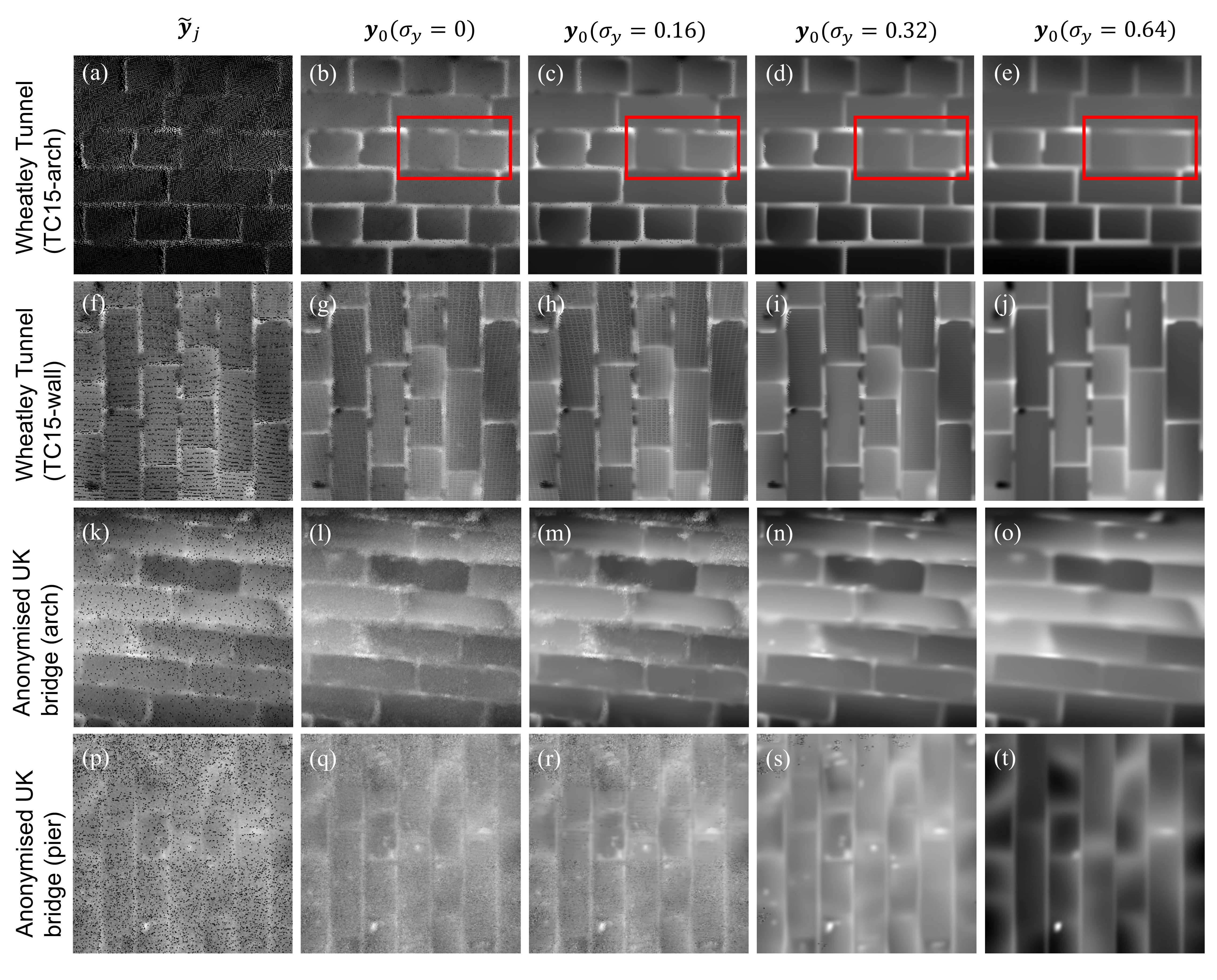}
    \caption{Effect of varying $\sigma_y \in \{0, 0.16, 0.32, 0.64\}$ on IR quality. (a)–(e) and (f)–(j) show $\tilde{\mathbf{y}}_j$ and $\mathbf{y}_0$ for the arch and walls of Wheatley Tunnel (TC15) (patch 1828 and patch 969, respectively). (k)–(o) and (p)–(t) show corresponding results for the arch and pier of the anonymised UK bridge (patch 703 and patch 145, respectively). The red boxes from (b)-(e) indicate the blurring of brick boundaries by increasing $\sigma_y$.}
    \label{fig: inpainting_noise}
\end{figure*}

Our evaluation of depth map restoration by InfraDiffusion is conducted in a qualitative manner, since quantitative GT supervision (dense depth maps) is unavailable. Instead, we demonstrate the effectiveness of IR through the semantic segmentation by leveraging zero-shot and prompt-based SAM. The IR results are demonstrated in Figure~\ref{fig: inpainting_arch}a-j and Figure~\ref{fig: inpainting_pier}a-j across the five cases: the three selected Wheatley Tunnel chunks (TC15, TC20, and TC25), the anonymised UK bridge, and the Hertford Viaduct, where $\sigma_y = 0.16$. The selected patch indices are summarised in Table~\ref{tab:patch_indices}, which correspond to Figures~\ref{fig: inpainting_arch}–\ref{fig: inpainting_noise}. 

The different orientations of $\tilde{\mathbf{y}}_j$ between arch and wall/pier arise from the default generatrix direction $\mathbf{v}_j$ (as shown in Figure~\ref{fig: patch_projection}) used during virtual camera projection. The different orientations of $\tilde{\mathbf{y}}_j$ between arch and wall/pier arise from the default generatrix direction $\mathbf{v}_j$ (as shown in Figure~\ref{fig: patch_projection}) used during virtual camera projection. While this choice influences the visual orientation and the density distribution of projected points, we observed that InfraDiffusion produces consistently plausible restorations across viewpoints given robust normal estimations.

By comparing results for the anonymised UK bridge (Figure~\ref{fig: inpainting_arch}d, e, i, and j) and Hertford Viaduct (Figure~\ref{fig: inpainting_pier}d, e, i, and j), the arch surfaces exhibit blurrier and less distinct brick boundaries than the pier surfaces. This discrepancy is mainly due to the TLS acquisition geometry discussed in Section~\ref{sec: wheatley_dataset}, where arches are scanned from greater distances than piers, resulting in lower point density and higher noise.

We further examine the influence of varied $\sigma_y$ on restoration quality in Figure~\ref{fig: inpainting_noise}a–t. Results are illustrated for both the Wheatley Tunnel (arch patch 1828 and wall patch 969) and the anonymised UK bridge (arch patch 703 and pier patch 145), which indicate that the assumed noise level in datasets strongly influences the denoising strength. For $\sigma_y=0$, the restored images still exhibit residual speckle and blurred mortar boundaries (see Figure~\ref{fig: inpainting_noise}b, g, l, and q). As $\sigma_y$ increases to 0.16, the background noise is effectively suppressed while brick boundaries remain clear and continuous, producing the most visually coherent results (Figure~\ref{fig: inpainting_noise}c, h, m, and r). At $\sigma_y=0.32$ and $\sigma_y=0.64$, the denoising starts to blur the brick edges within the red boxes in Figure~\ref{fig: inpainting_noise}b–e progressively, and fine boundary details are lost despite smoother textures.  

Based on this observation, we select $\sigma_y=0.16$ for all subsequent experiments. This choice represents a practical balance between noise suppression and boundary preservation. We note, however, that the optimal value is empirically and visually determined and may vary across datasets due to differences in point density, registration errors, and sensor noise. For consistency and ease of comparison, we apply the same $\sigma_y$ across all cases in this study.

\begin{figure}[t!]
    \centering
    \includegraphics[width=\linewidth]{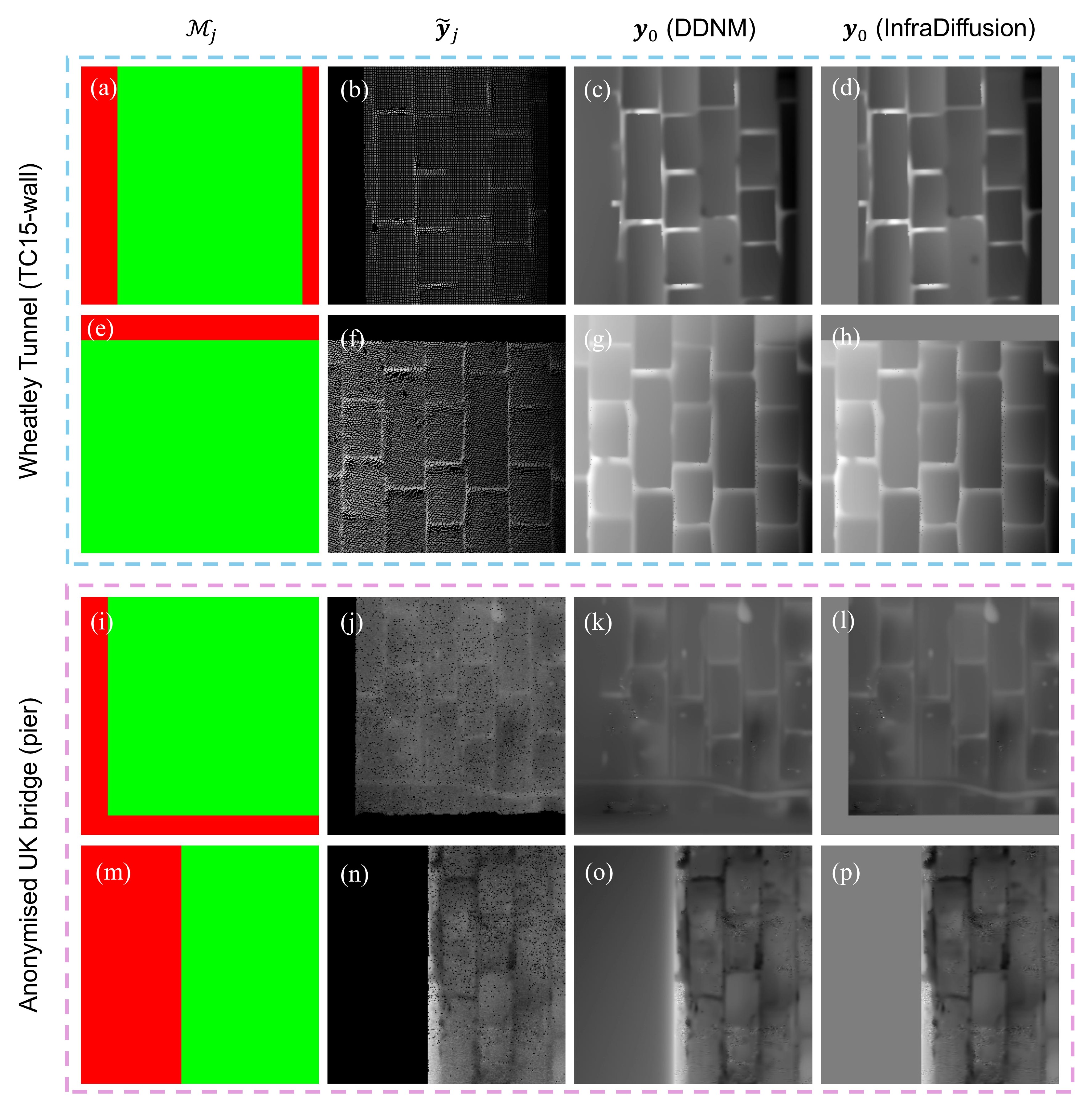}
    \caption{Comparison between DDNM and InfraDiffusion on wall/pier patches. (a)–(d) and (e)–(h) represent Wheatley Tunnel (TC15) wall patches 897 and 1194, respectively. (i)–(l) and (m)–(p) correspond to anonymised UK bridge pier patches 265 and 479. For each patch, the four columns show: boundary mask $\mathcal{M}_j$ (colours are defined in Figure~\ref{fig: boundary_mask}), projected depth map $\tilde{\mathbf{y}}_j$, restoration $\mathbf{y}_0$ by DDNM, and restoration by InfraDiffusion.} 
    \label{fig: ablation_test}
\end{figure}

\subsection{Ablation test on InfraDiffusion}

To demonstrate the effectiveness of the proposed InfraDiffusion conditioned on $\mathcal{M}_j$, we compare $\mathbf{y}_0$ from the vanilla DDNM and our InfraDiffusion framework. All selected patches are taken from walls/piers, where structural boundaries are common, such as wall/pier edges or regions connected to the ground. Figure~\ref{fig: ablation_test}a-p shows four representative cases: Wheatley Tunnel (patches 897 and 1194 of TC15) and the anonymised UK bridge (patches 265 and 476). For each case, we present the $\mathcal{M}_j$, $\tilde{\mathbf{y}}_j$, and the corresponding IR results (e.g., $\mathbf{y}_0$) using DDNM and InfraDiffusion.

The results indicate that DDNM, while achieving the same results in foreground regions (defined in Figure~\ref{fig: boundary_mask}), often produces spurious structures beyond the valid projection region (e.g., background regions). This is particularly evident in Figures~\ref{fig: ablation_test}g and \ref{fig: ablation_test}k, where non-existent bricks are generated outside the wall/pier boundary. By contrast, InfraDiffusion constrains restoration strictly within the masked region, preserving true edges and avoiding hallucinated patterns (Figures~\ref{fig: ablation_test}d, h, l, and p). Across all cases, InfraDiffusion yields more physically consistent restorations by preserving the original spatial structures of point clouds.

\begin{figure*}[t!]
    \centering
    \includegraphics[width=\linewidth]{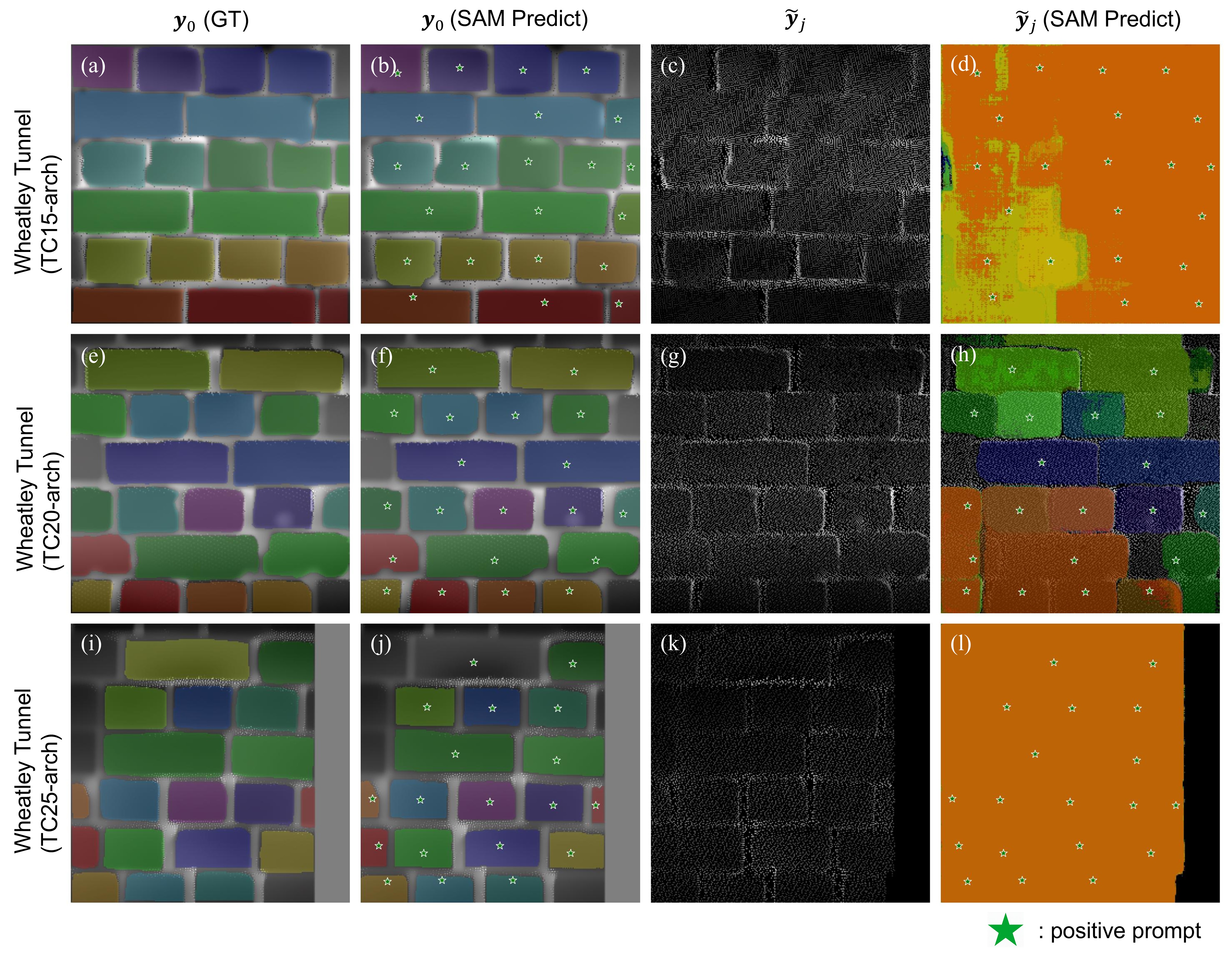}
    \caption{Zero-shot segmentation of arch patches from Wheatley Tunnel using SAM with ViT-H. (a)–(d) show GT ($\mathbf{y}_0$), SAM predictions on $\mathbf{y}_0$, $\tilde{\mathbf{y}}_j$, and SAM predictions on $\tilde{\mathbf{y}}_j$ for test chunk 15. (e)–(h) and (i)–(l) correspond to test chunks 20 and 25, respectively. Green pentagrams indicate positive prompts.}
    \label{fig: SAM_tunnel_arch}
\end{figure*}

\subsection{Zero-shot segmentation with SAM}

We further evaluate the effectiveness of the InfraDiffusion for downstream segmentation tasks using SAM \citep{ye2024sam}. Specifically, we adopt the ViT-H SAM model weights to generate segmentation masks for masonry brick boundaries. GT annotations are created only for bricks with clear and well-defined boundaries, while instances with blurred, indistinct, and incomplete outlines are omitted for rigorous performance. The labelling is performed on the sampled patches introduced in the Dataset section.

We adopt prompt-based SAM rather than training/fine-tuning DL-based segmentation methods such as YOLO \citep{khanam2024yolov11}, as the latter requires large labelled datasets that are often unavailable for masonry structures. In contrast, SAM, which supports zero-shot operation, can better reflect realistic annotation scenarios, where practitioners rely on AI-assisted tools by providing prompts (e.g., X-AnyLabeling \citep{X-AnyLabeling}) to speed up labelling while retaining user control.

\begin{figure*}[t!]
    \centering
    \includegraphics[width=\linewidth]{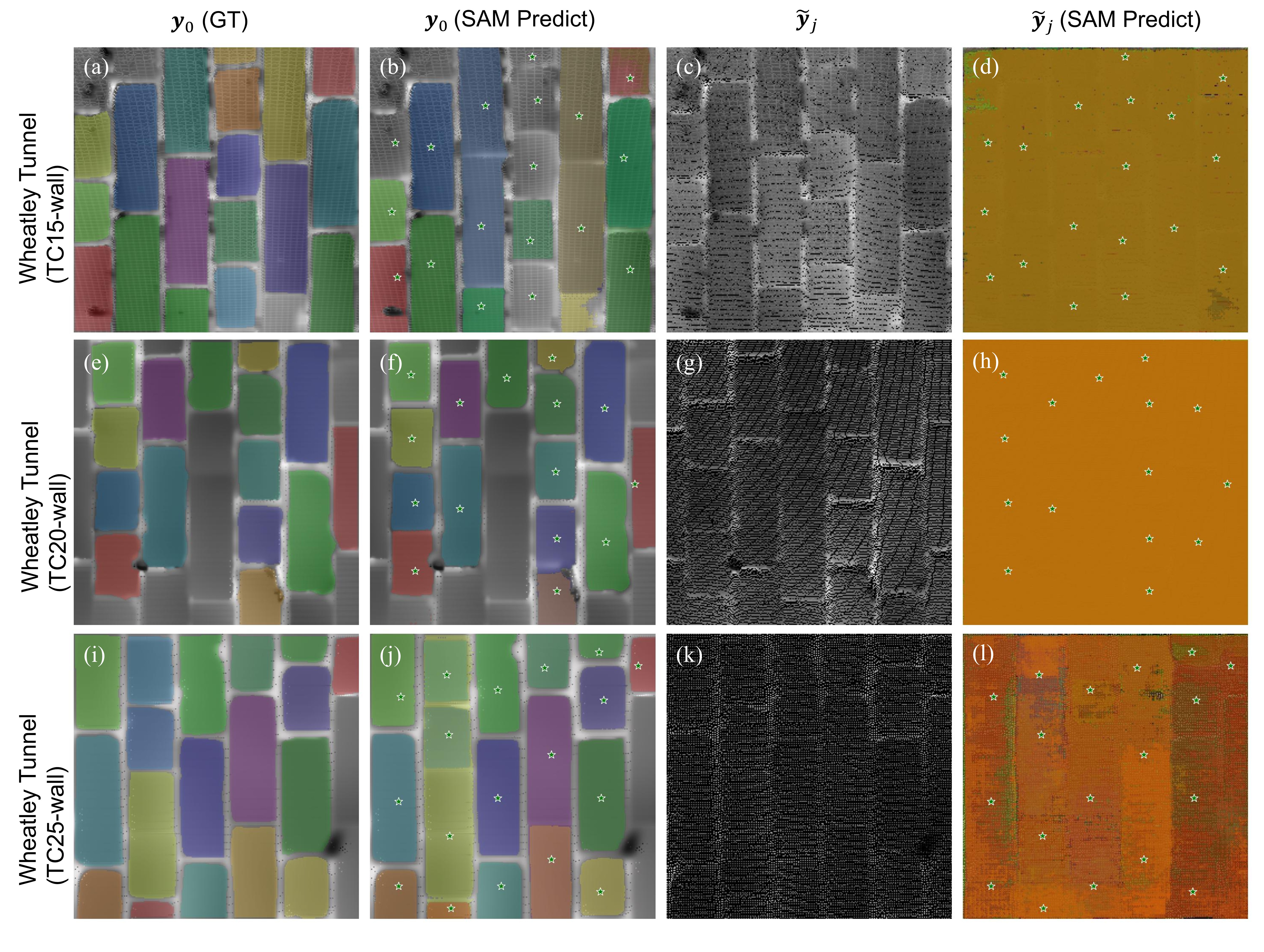}
    \caption{Zero-shot segmentation of wall patches from Wheatley Tunnel using SAM with ViT-H. (a)–(d) represent GT ($\mathbf{y}_0$), SAM predictions on $\mathbf{y}_0$, $\tilde{\mathbf{y}}_j$, and SAM predictions on $\tilde{\mathbf{y}}_j$ of test chunk 15. (e)–(h) and (i)–(l) correspond to test chunks 20 and 25. Green pentagrams indicate positive prompts.}
    \label{fig: SAM_tunnel_pier}
\end{figure*}

The experiments use the same patch indices listed in Table~\ref{tab:patch_indices} of the depth map restoration. To simulate the interaction between humans and AI-assisted annotation tools, prompts are placed at representative brick regions. A positive prompt is generated by averaging the centroids of GT polygons, while the negative prompt is a randomly sampled point outside the annotated regions. Negative prompts are used only for the masonry bridges to mitigate the influence of lower data quality, but not for the tunnel experiments (only positive prompts are used). Since random negative prompts can scatter across the image plane and introduce confusion, they are omitted from the visualisations but retained in the quantitative analysis.

\begin{table}[t!]
\centering
\caption{mIoU values of SAM segmentation on $\tilde{\mathbf{y}}_j$ and $\mathbf{y}_0$ using different prompting strategies.}
\label{tab:sam_miou}
\resizebox{\textwidth}{!}{%
\begin{tabular}{lccccc}
\toprule
& \makecell{Wheatley \\ Tunnel (TC15)} & \makecell{Wheatley \\ Tunnel (TC20)} & \makecell{Wheatley \\ Tunnel (TC20)} & \makecell{Anonymised\\UK bridge} & \makecell{Hertford\\viaduct} \\
\midrule
$\tilde{\mathbf{y}}_j$ (1 pos) & 0.064 & 0.216 & 0.181 & 0.047 & 0.064 \\
$\tilde{\mathbf{y}}_j$ (1 pos + 1 neg) & / & / & / & 0.049 & 0.074 \\
$\mathbf{y}_0$ (1 pos) & 0.708 & 0.875 & 0.780 & 0.436 & 0.708 \\
$\mathbf{y}_0$ (1 pos + 1 neg) & / & / & / & 0.460 & 0.729 \\
\bottomrule
\end{tabular}
}
\end{table}

Figure~\ref{fig: SAM_tunnel_arch}a–l and Figure~\ref{fig: SAM_tunnel_pier}a–l illustrate zero-shot segmentation results on the arch and wall components of Wheatley Tunnel (TC15, TC20, and TC25) using SAM (ViT-H). As shown in Figure~\ref{fig: SAM_tunnel_arch}a, e, and i and Figure~\ref{fig: SAM_tunnel_pier}a, e, and i, the transparent coloured overlays denote the GT masks of masonry bricks. The corresponding SAM predictions on $\mathbf{y}_0$ with positive prompts only are shown in Figure~\ref{fig: SAM_tunnel_arch}b, f, and j and Figure~\ref{fig: SAM_tunnel_pier}b, f, and j. For visual clarity, the colours of the predicted masks are matched to the GT labels. Only brick instances with an Intersection-over-Union (IoU) score greater than 0.3 are displayed to prevent overlapping of incorrect masks. The IoU of a single masonry brick is defined as:

\begin{equation}
\text{IoU} = \frac{|\mathcal{M}_{\text{GT}} \cap \mathcal{M}_{\text{Pred}}|}{|\mathcal{M}_{\text{GT}} \cup \mathcal{M}_{\text{Pred}}|}
\end{equation}
where $\mathcal{M}_{\text{GT}}$ and $\mathcal{M}_{\text{Pred}}$ denote the GT and predicted masks of brick instances, respectively. 

In contrast, Figure~\ref{fig: SAM_tunnel_arch}d, h, and l and Figure~\ref{fig: SAM_tunnel_pier}d, h, and l present SAM predictions on $\tilde{\mathbf{y}}_j$, visualised using the same positive prompts. To make the comparison consistent, we show the same set of masks selected for $\mathbf{y}_0$, but without IoU-based filtering. The predictions on $\tilde{\mathbf{y}}_j$ collapse into large and overlapping masks that fail to delineate brick boundaries.

\begin{figure*}[t!]
    \centering
    \includegraphics[width=\linewidth]{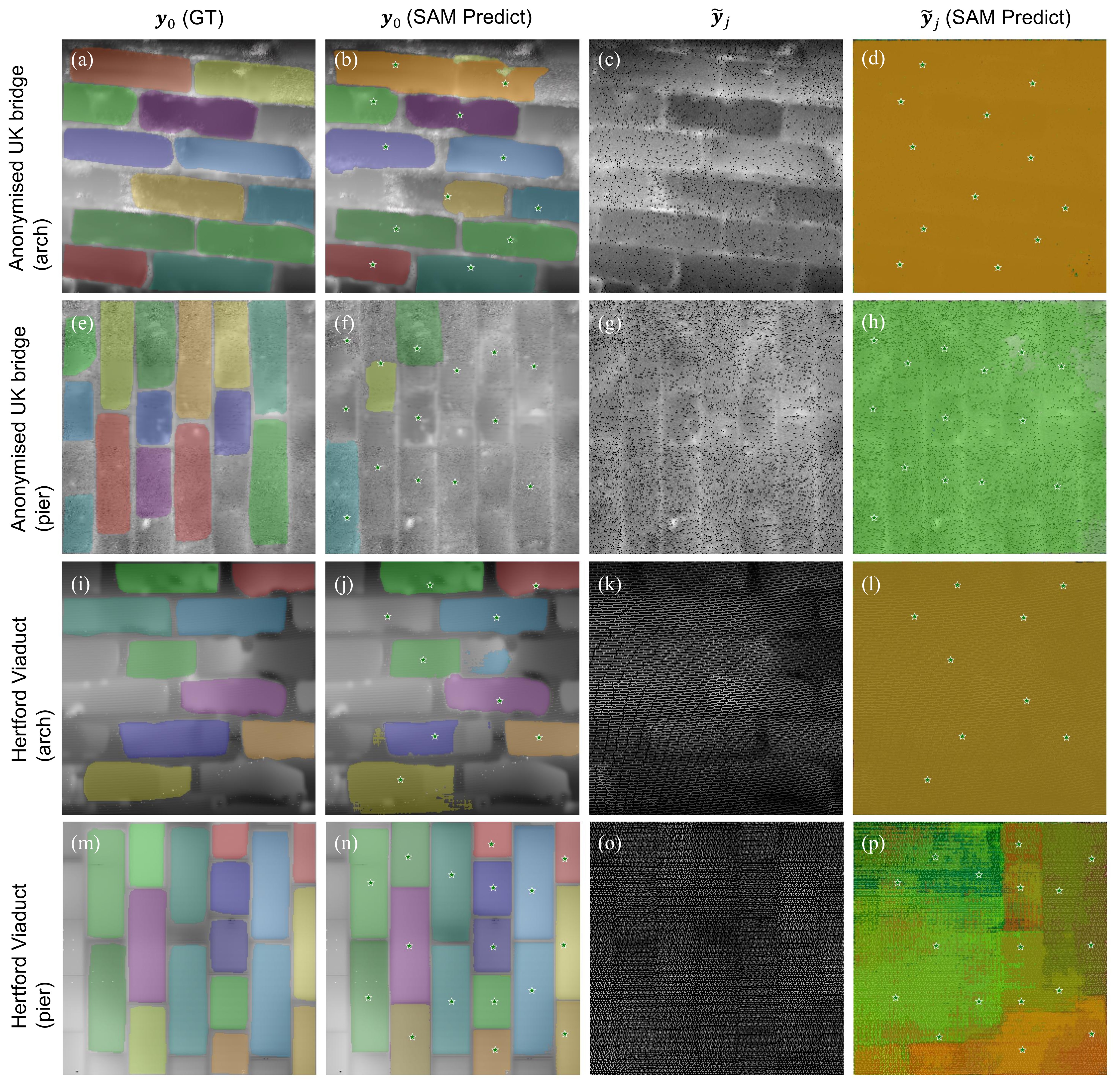}
    \caption{Zero-shot segmentation of masonry bridge patches using SAM with ViT-H by providing both positive and negative prompts. (a)–(d) show results for the arch of the anonymised UK bridge, (e)–(h) for the pier of the anonymised UK bridge, (i)–(l) for the arch of the Hertford Viaduct, and (m)–(p) for the pier of the Hertford Viaduct.}
    \label{fig: SAM_bridge}
\end{figure*}

To quantify segmentation performance, we compute the mean Intersection-over-Union ($\text{mIoU}$) by first evaluating the IoU for each annotated brick instance, then averaging across all bricks within an infrastructure, and finally averaging across all patches in the dataset. As summarised in Table~\ref{tab:sam_miou}, the predictions on $\mathbf{y}_0$ achieve substantially higher scores than those on $\tilde{\mathbf{y}}_j$. For TC15, TC20, and TC25 of Wheatley Tunnel, the mIoU improves from 0.064, 0.216, and 0.181 on $\tilde{\mathbf{y}}_j$ to 0.708, 0.875, and 0.780 on $\mathbf{y}_0$, respectively, which demonstrates the effectiveness of InfraDiffusion in enabling accurate brick-level segmentation.

For the bridge datasets, we evaluate two prompting strategies. The first uses a single positive prompt, consistent with the tunnel experiments, while the second combines a positive–negative prompt pair to increase robustness as the masonry datasets are noisier. As shown in Table~\ref{tab:sam_miou}, using a single positive prompt yields mIoU scores of 0.436 for the anonymised UK bridge and 0.708 for the Hertford viaduct. Adding a negative prompt raises the mIoU to 0.460 and 0.729, respectively. 

The anonymised UK bridge presents a more challenging case due to the lower quality of $\tilde{\mathbf{y}}_j$ (Figure~\ref{fig: SAM_bridge}c and g), in that the boundaries are more blurred compared to the Wheatley Tunnel shown in Figure~\ref{fig: SAM_tunnel_arch}e and g). Therefore, SAM predictions fail to capture meaningful masonry boundaries and degenerate into large spreading masks (Figure~\ref{fig: SAM_bridge}d and h). By contrast, InfraDiffusion substantially improves segmentation on $\mathbf{y}_0$ (Figure~\ref{fig: SAM_bridge}b and f), where SAM (guided by positive–negative prompts) recovers coherent brick-level boundaries. The improvement is reflected in the quantitative results, where the mIoU increases from 0.049 on $\tilde{\mathbf{y}}_j$ to 0.460 on $\mathbf{y}_0$ (Table~\ref{tab:sam_miou}).

For the Hertford Viaduct, both arch and pier patches benefit from InfraDiffusion, as shown in Figure~\ref{fig: SAM_bridge}i–l and m–p, respectively. However, the improvement is more evident for the pier regions. $\mathbf{y}_0$ of the pier yields sharper and more continuous brick boundaries, enabling SAM to recover detailed masks, whilst arch predictions are still influenced by blurred boundaries even after InfraDiffusion, as shown in Figure~\ref{fig: SAM_bridge}i and j. 

Such a discrepancy between arch and pier performance is primarily attributable to the relative distance between the TLS and the structural component. As shown in Figure~\ref{fig: bridge_segmentation}, the height of the Hertford Viaduct arch is 12.6 m, which results in noisier $\tilde{\mathbf{y}}_j$. In contrast, the piers are located at a lateral distance of approximately half the bridge width (4.95 m) from the scanner when positioned at the centreline, leading to denser and cleaner projections. This trend is not observed in the anonymised UK bridge, where a smaller height of 3.7 m ensures that both arches and piers remain within a similar scanning range, which results in more uniform (albeit consistently lower-quality) inputs. 

As shown in Table~\ref{tab:sam_miou}, the mIoU on the Hertford Viaduct improves from 0.064 on $\tilde{\mathbf{y}}_j$ to 0.708 on $\mathbf{y}_0$ when using a single positive prompt. With the positive–negative prompting strategy, the score further rises to 0.729 (Table~\ref{tab:sam_miou}). These results confirm the effectiveness of InfraDiffusion in restoring depth map quality and demonstrate its effectiveness and robustness in supporting downstream zero-shot segmentation tasks.

\section*{Conclusions}

This work introduced \textbf{InfraDiffusion}, a zero-shot diffusion framework that restored sparse and noisy depth maps derived from masonry point clouds, enabling zero-shot brick-level image segmentation for structural assessment. The pipeline first applied a virtual camera projection to systematically generate depth maps from large-scale masonry infrastructure. It then adapted the vanilla DDNM framework with boundary masking to restore these maps, addressing the boundary effects of point cloud projection in civil engineering applications. By constraining the restoration to valid regions only, InfraDiffusion prevented spurious content outside structural edges. InfraDiffusion relied solely on pre-trained DDPMs to provide strong generative priors, operating entirely in a zero-shot setting without requiring task-specific training or ground-truth supervision.

Extensive experiments were conducted on three representative chunks of the Wheatley Tunnel (TC15, TC20, and TC25 with 100 patches) and two masonry bridges (an anonymised UK bridge with 60 patches and the Hertford Viaduct with 140 patches). InfraDiffusion consistently produced clean and geometrically coherent depth maps across these datasets. The ablation study further demonstrated the effectiveness of InfraDiffusion conditioned on extra boundary masks. While vanilla DDNM produces the same results in foreground regions, it also generates spurious bricks outside structural edges when applied to projected depth maps. InfraDiffusion suppresses such artefacts by constraining restoration to masked regions, yielding more physically consistent outputs for infrastructure point clouds.  

Extensive experiments were conducted on three representative chunks of the Wheatley Tunnel (TC15, TC20, and TC25 with 100 patches) and two masonry bridges (an anonymised UK bridge with 60 patches and the Hertford Viaduct with 140 patches). InfraDiffusion consistently produced clean and geometrically coherent depth maps across these datasets. The ablation study further demonstrated the effectiveness of conditioning DDNM with boundary masks. While vanilla DDNM produced comparable results in foreground regions, it also generated spurious bricks outside structural edges when applied to projected depth maps. InfraDiffusion suppressed such artefacts by constraining restoration to masked regions, yielding more physically consistent outputs for infrastructure point clouds. 

Quantitative validation was carried out through zero-shot semantic segmentation using SAM. With only a single positive prompt inside bricks, InfraDiffusion significantly improved segmentation performance: across all datasets, mIoU scores rose from as low as 0.064 on Wheatley Tunnel TC15 to 0.708 after restoration. Additional robustness was achieved by introducing a negative prompt for the two noisier bridge datasets, where mIoU further improved from 0.436 to 0.460 on the anonymised UK bridge and from 0.708 to 0.729 on the Hertford Viaduct. These results showed that InfraDiffusion effectively transformed noisy depth maps into representations that downstream segmentation models could exploit.

Our results demonstrated the robustness and effectiveness of InfraDiffusion across both tunnels and bridges. They highlighted the potential of diffusion models for civil engineering: despite being trained solely on natural images, pre-trained diffusion models transferred effectively to infrastructure depth maps because their strong generative priors captured local pixel dependencies that were equally relevant in masonry. This pointed to promising avenues for applying diffusion models to other structural assessment tasks where sparse and noisy sensing data must be transformed into complete and analysable representations.

\textbf{Limitations and future work}  While InfraDiffusion demonstrates robustness and effectiveness, several limitations remain. First, although DDIM accelerates sampling compared to DDPM, the restoration process is still computationally intensive when applied to large numbers of structural patches, limiting scalability for large-scale masonry infrastructures. 

Second, the current framework is not end-to-end.  The IR of InfraDiffusion and brick-level segmentation are performed sequentially, whereas future work should aim to integrate these steps into a unified pipeline for improved efficiency and consistency.  

Third, our method restores only geometric depth information, without considering generating synthetic RGB colour, which is often unavailable in LiDAR scans taken from low-light conditions. Incorporating colour restoration could further enhance the point cloud quality and support downstream tasks such as defect detection.  

\section*{Data availability statement} 

The data and the code are now available at \url{https://github.com/Jingyixiong/InfraDiffusion-official-implement}.

\bibliographystyle{elsarticle-harv}     
\bibliography{update_patchcore_citations}

\end{document}